\pdfoutput=1
\documentclass[11pt]{article}
\usepackage[table]{xcolor}
\usepackage[preprint]{acl}

\usepackage{times}
\usepackage{latexsym}
\usepackage{arydshln}
\usepackage{arydshln}
\usepackage[T1]{fontenc}
\usepackage[utf8]{inputenc}

\usepackage{soul}
\usepackage{subfigure}
\usepackage{subfiles}
\usepackage{microtype}

\usepackage{inconsolata}
\usepackage{listings}
\usepackage{graphicx}
\usepackage{subfigure}
\usepackage{algorithm}
\usepackage{algorithmic}
\usepackage{amsmath}
\usepackage{amssymb}
\usepackage{mathtools}
\usepackage{amsthm}
\usepackage{enumitem}
\usepackage{comment}
\usepackage{subcaption}

\usepackage{makecell}
\title{OmniReflect: Discovering Transferable Constitutions\\ for LLM agents via Neuro-Symbolic Reflections}
\author{
\textbf{Manasa Bharadwaj }  \hspace{0.2cm}
 \textbf{Nikhil Verma} \hspace{0.2cm}
 \textbf{Kevin Ferreira} \\
   LG Electronics, Toronto AI Lab \\ 
  \texttt{\{manasa.bharadwaj,nikhil.verma,kevin.ferreira\}@lge.com}
}

\begin{document}
\maketitle

\begin{abstract}
Efforts to improve Large Language Model (LLM) agent performance on complex tasks have largely focused on fine-tuning and iterative self-correction. 
However, these approaches often lack generalizable mechanisms for long-term learning and remain inefficient in dynamic environments.
We introduce OmniReflect, a hierarchical, reflection-driven framework that constructs a constitution, a compact set of guiding principles distilled from task experiences, to enhance the effectiveness and efficiency of an LLM agent.
% We introduce OmniReflect, a hierarchical reflection-driven framework that constructs constitutions, compact sets of guiding principles distilled from past task experiences, to enhance the effectiveness and efficiency of LLM agents.
%
OmniReflect operates in two modes: Self-sustaining, where a single agent periodically curates its own reflections during task execution, and Co-operative, where a meta-advisor derives a constitution from a small calibration set to guide another agent.
% OmniReflect operates in two modes: Self-sustaining, where a single agent periodically generates and summarizes its own reflections during task execution, and Co-operative, where a meta-advisor derives constitutions from a small calibration set to guide another agent.
% OmniReflect operates in two modes: Self-sustaining, where a single agent periodically generates and summarizes its own reflections, and Co-operative, where a meta-advisor generates constitutions from a small calibration set for another agent to use.
%
% We use Neural, Symbolic, and Neuro-Symbolic methods to build these constitutional principles, balancing flexibility with computational efficiency.
To construct these constitutional principles, we employ Neural, Symbolic, and Neuro-Symbolic techniques, offering a balance between contextual adaptability and computational efficiency.
Empirical results averaged across models show major improvements in task success, with absolute gains of +10.3\% on ALFWorld, +23.8\% on BabyAI, and +8.3\% on PDDL in the Self-sustaining mode.
% Empirical results demonstrate that OmniReflect consistently improves task success across agentic benchmarks, achieving absolute gains of +10.3\% (ALFWorld), +23.8\% (BabyAI), and +8.3\% (PDDL) in the Self-sustaining mode.
%
% Comparable improvements are observed in the Co-operative mode as well.
Similar gains are seen in the Co-operative mode, where a lightweight Qwen3-4B ReAct agent outperforms all Reflexion baselines on BabyAI. % highlighting the collaborative potential of OmniReflect.
These findings highlight the robustness and effectiveness of OmniReflect across environments and backbones.
% These findings establish OmniReflect as a robust framework for reflection-driven learning, demonstrating consistent performance gains across environments and model scales, and paving the way for scalable, adaptable, and efficient LLM agents.
\end{abstract}

\section{Introduction}\label{intro}

\begin{figure}[h]
    \centering
    \includegraphics[width=0.9\linewidth]{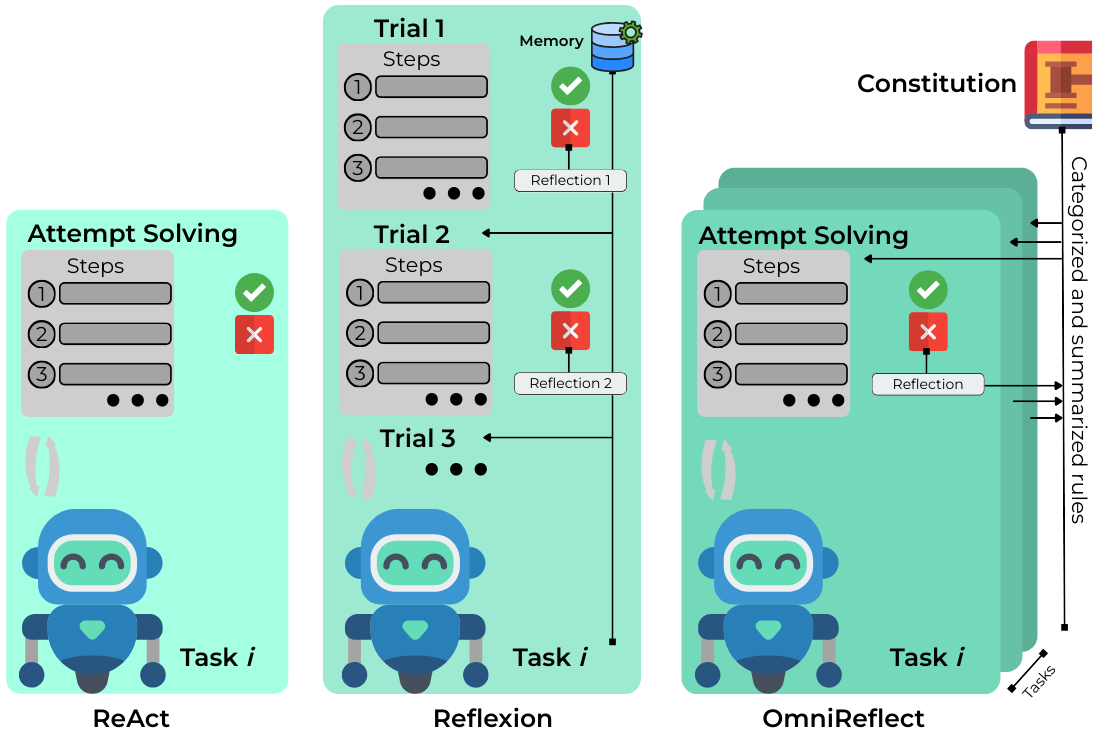}
    \caption{Existing strategies such as ReAct~\cite{yao2022react} rely on step-by-step reasoning within a single trajectory, while Reflexion~\cite{shinn2024reflexion} enhances performance through iterative retries guided by self-critiques. In contrast, OmniReflect maintains categorized constitution rules (summarized reflections) that guide task-solving by collating knowledge over time.%in a step-wise manner. These rules persist across tasks within the environment.
    }
    \label{fig:omnireflect_comparison}
\end{figure}

\begin{figure*}[th]
    \centering
    \includegraphics[width=0.9\textwidth]{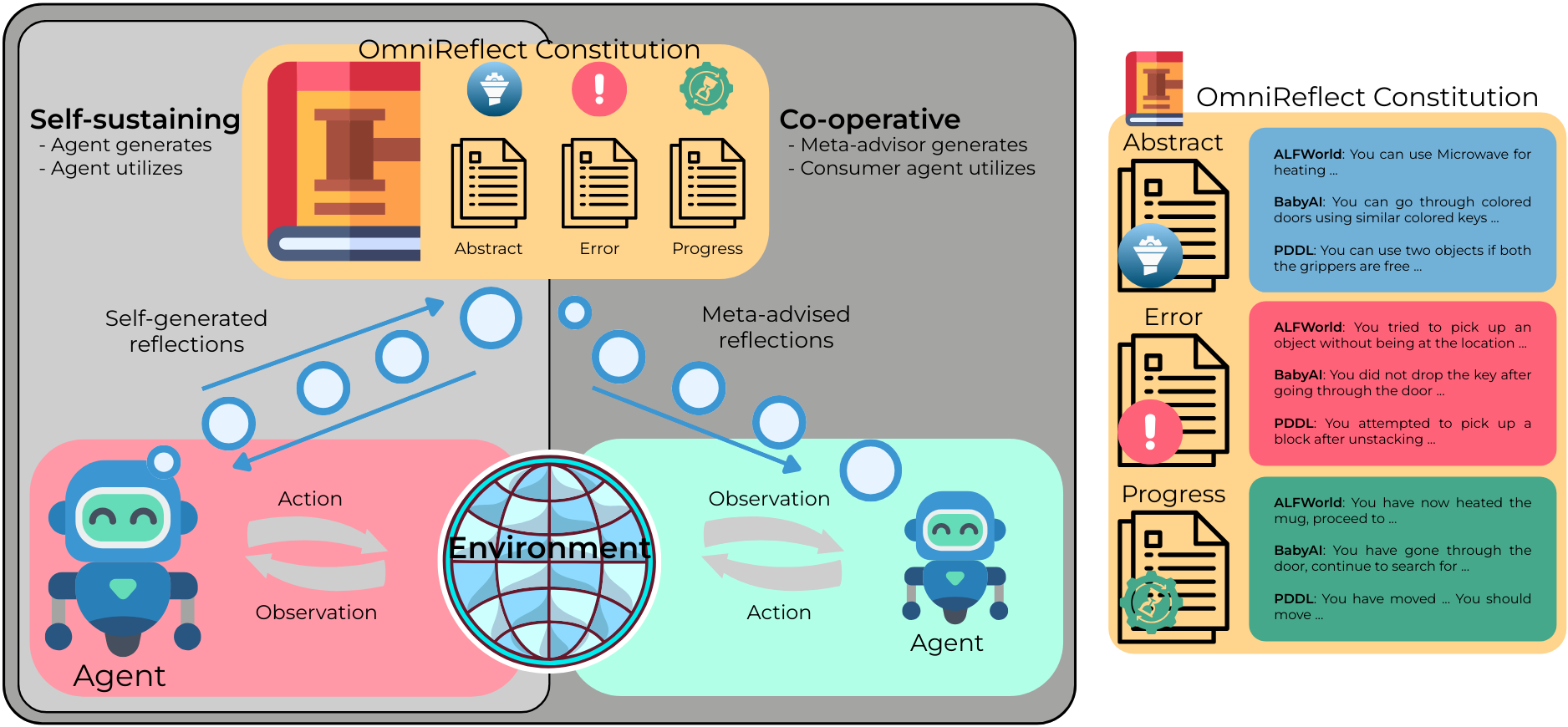}
    \caption{The OmniReflect framework operates in two reflective modes: (1) Self-Sustaining: where the LLM agent periodically generates, updates and uses the constitution; and (2) Co-operative: where a Meta-advisor derives constitution rules from a calibration set to guide another agent. These constitutional rules encode task level (low) and environment level (high), knowledge for effective task completion. Examples are shown on the right.}
    \label{fig:omnireflect_overview}
\end{figure*}

%-------------------------------
% Paragraph 1
%-------------------------------
% Foundational Large Language Models (LLMs) have been increasingly used as autonomous agents to learn about an environment and associated tools to perform often compound and complex tasks for the user \cite{yang2023auto, nakajima2023babyagi, shen2024hugginggpt, ahn2022can, huang2022inner}.
% %
% Despite this progress, studies continue to underscore the limitations of LLM-based agents, particularly in reasoning, planning, and continual learning \cite{jain2024ragmodulo, self-correct-yet}.
%-------------
Foundational Large Language Models (LLMs) are increasingly being deployed as autonomous agents to explore environments and utilize tools for executing complex, often compound tasks on behalf of users \cite{shen2024hugginggpt, yang2023auto, nakajima2023babyagi}. 
Despite these advancements, recent studies continue to highlight key limitations of LLM-based agents, particularly in reasoning, planning, and continual learning \cite{jain2024ragmodulo, self-correct-yet}.

Two major strategies are commonly employed to mitigate these limitations:
\begin{enumerate}
    \item \textbf{Fine-tuning}, typically performed on large-scale, environment-specific datasets using supervised or reinforcement learning approaches \cite{guo2025deepseek, deng2024mind2web, shridhar2020alfworld}, which is computationally expensive and often lacks scalability.
    \item \textbf{Reasoning and self-correction}, where an LLM agent is prompted step-by-step to reason about its actions and revise its course if the task remains incomplete \cite{shinn2024reflexion, madaan2023self, xi2023selfpolish}.
\end{enumerate}

Self-correcting methods usually involve analyzing and critiquing failed attempts to extract task-specific feedback~(called reflection) that can guide future trials \cite{shinn2024reflexion, madaan2023self}.
%
% Although these approaches enhance task performance, they often fall short in achieving generalization in an environment \cite{xie2024travelplanner}.
Although these approaches enhance task-level performance, they often fall short in achieving abstraction and generalization in an environment \cite{xie2024travelplanner}.

%-------------------------------
% Paragraph 3
%-------------------------------
% We introduce \textbf{OmniReflect}, an advanced hierarchical self-reflection framework designed to adaptively curate evolving constitutions at both the environmental and task levels, in tandem with task execution. 
% By intricately fusing self-reflection with task execution, OmniReflect not only improves task completion and efficiency but also fosters the generation of re-usable knowledge.
% To preserve the integrity of the constitutions, such as avoiding redundancies, OmniReflect conducts regular summarizations, converting disorganized episodic experiences into well-structured, reusable long-term memory.
%-----------
We introduce OmniReflect, a hierarchical reflection-based framework that adaptively curates an evolving \textit{constitution}, a set of categorized and summarized reflections, to guide task-solving by accumulating knowledge over time.
The constitution serves as structured memory, distilled from prior and current task experiences within an environment.
By tightly integrating reflections with task execution, OmniReflect not only enables efficient task completion but also facilitates the creation of reusable knowledge.
To ensure the continued utility and coherence of the constitution~(e.g., reducing redundancy), OmniReflect performs periodic summarization, transforming unstructured episodic traces into well-organized rule book.
Figure~\ref{fig:omnireflect_comparison} provides an overview of the OmniReflect framework and highlights key differences compared to existing approaches.

%-------------------------------
% Paragraph 4
%-------------------------------
% OmniReflect can be deployed in two distinct modes: 1) Self-sustaining Agent and 2) Meta-Advisor. 
% The Self-sustaining Agent mode, depicted in Figure \ref{fig:omnireflect_overview} (left), has been previously detailed. 
% In order to overcome the cold-start problem for self-reflection based agents, which predominantly impacts simple agents, OmniReflect can also be used as a meta-advisor, as illustrated in Figure \ref{fig:omnireflect_overview} (right), which we refer to as the cooperative mode.
% In this mode, constitutions discovered and collated by an OmniReflect agent through the utilization of a calibration set can be effectively embedded into the prompts of lightweight downstream agents. 
% This integration results in significant performance improvements with negligible additional computational overhead. Furthermore, the synergistic combination of these two modes has the potential to develop a robust system that acquires foundational knowledge during post-training calibration and continuously refines it during test-time, akin to the process of human learning.
%-------------
OmniReflect can be used in two distinct modes: 1) Self-sustaining and 2) Co-operative.
Figure~\ref{fig:omnireflect_overview} illustrates how both modes generate and utilize constitution, structured reflections accumulated across the environment.
In the Self-sustaining mode, the same LLM agent periodically generates and reuses summarized reflections, curating them at fixed step intervals.
In the Co-operative mode, a Meta-advisor agent uses a small calibration set of tasks to generate constitution rules through reflections, which are then consumed by a separate task-solving agent.
This modular integration leads to consistent performance improvements with minimal computational overhead.
Furthermore, the synergistic integration of these two modes creates a robust system that acquires foundational knowledge during post-training calibration and incrementally refines it at test time, complementing both fine-tuning and self-correction strategies.
We leverage Neural, Symbolic, and Neuro-Symbolic techniques to construct a constitution.
%from accumulated reflections.
%
Neural methods offer rich, context-aware feedback but are often computationally intensive. 
Symbolic methods, while efficient and interpretable, tend to lack flexibility.
Our Neuro-Symbolic approach strikes a balance, combining the adaptability of neural models with the structural efficiency of symbolic reasoning to enable scalable and cost-effective constitution generation.
We empirically validate this design across diverse agentic benchmarks and LLMs, demonstrating that OmniReflect consistently outperforms strong baselines while solving tasks in fewer steps, highlighting both its effectiveness and efficiency.

%-------------------------------
% Paragraph end - Hi NIk! - Hello dear
%-------------------------------
\noindent Our key contributions are as follows:

\begin{itemize}
    % \item \hl{We introduce the OmniReflect framework, which illustrates the impact of hierarchical self-reflection at both task and environmental levels when coupled with the processes of knowledge creation, retention, and dissemination, in conjunction with task execution can improve task performance.}
    % \item We propose OmniReflect, a hierarchical reflection-driven agentic framework that enables LLMs to accumulate reusable knowledge across environment through adaptive creation and summarization of constitutions that improve overrall task accuracy and efficiency.
    %\item We propose OmniReflect, a hierarchical reflection-driven agentic framework that enables LLMs to adaptively summarize task level reflections and accumulate reusable knowledge across environment.
    % \item We propose OmniReflect, a hierarchical, reflection-driven agent framework that enables LLMs to accumulate reusable knowledge by adaptively creating and summarizing constitutions in parallel with task execution, thereby enhancing reasoning efficiency and overall agent performance.
    % \item We propose OmniReflect, a hierarchical, reflection-driven framework that enables LLMs to accumulate reusable knowledge by adaptively creating and summarizing reflections into a constitution in parallel with task execution, thereby enhancing reasoning efficiency and overall agent performance.
    \item We propose OmniReflect, a hierarchical, reflection-driven framework that enables accumulation of reusable knowledge by adaptively creating and summarizing reflections into a constitution in parallel with task execution, thereby enhancing reasoning efficiency and overall agent performance.
    
    \item We demonstrate the robustness of OmniReflect{'}s constitution-building in two modes~(Self-sustaining and Co-operative) showing strong performance even with minimal calibration data.

    \item We evaluate OmniReflect across diverse agentic benchmarks (ALFWorld, BabyAI, and PDDL) using multiple LLM backbones, consistently outperforming competitive baselines in task success and efficiency.

    % \item In this paper, we present the OmniReflect framework to demonstrate that hierarchical self-reflection, at task and environment levels, leading to knowledge creation, retention and sharing along side task completion. 
    % \item We present OmniReflect, a hierarchical self-reflection driven framework designed to produce re-usable and shared, environment and task-level constitutions that can improve abstract learning and task-completion across agents.
    
    % \item We assess the performance of OmniReflect through three distinct agentic benchmarks: ALFWorld, BabyAI, and PDDL, and demonstrate substantial improvements as both a self-sustaining agent and as a meta-advisor, consistently across multiple LLM backbones.
    % % agents to act both in a self-sustaining mode and as a co-operative advisor for other simpler agents.
    % \item We empirically validate the robustness and transferability of the constitutions formulated by OmniReflect agents, demonstrating their efficacy even with minimal calibration data.
\end{itemize}

\section{OmniReflect}\label{omnireflect}
% Our modular framework for multi-dimensional hierarchical reflection, OmniReflect, is illustrated in Figure~\ref{fig:omnireflect_overview}, with the detailed workflow described in Algorithm~\ref{alg:omnireflect}.
The core design of our adaptable reflection framework, OmniReflect, is shown in Figure~\ref{fig:omnireflect_overview}, with step-by-step operations described in Algorithm~\ref{alg:omnireflect}.

% OmniReflect is designed to solve tasks within a dataset $D$ by leveraging an LLM agent $\mathcal{L}$ that alternates between two complementary phases: (1) \textsc{Action} ($\mathcal{L}_{\text{a}}$) and (2) \textsc{Reflection} ($\mathcal{L}_{\text{ref}}$).  
% %
% In the \textsc{Action} phase, the agent generates an action conditioned on the task description, current trajectory, and constitution ($OmniC$) curated upto the previous \textsc{Reflection} phase.  
% %
% In the \textsc{Reflection} phase, the agent generates both task-agnostic long-term memories (shared at the environment level) and task-specific progress reflections that serve as short-term knowledge to guide ongoing task completion.  
% %
% Section~\ref{sec:ref_types} provides a more detailed discussion of each reflection type.
% %
% All of the reflections are generated during $\mathcal{L}_{\text{ref}}$ phase, based on the current task description, observed trajectory, and historical constitution. 
%------------------
OmniReflect is designed to solve tasks from a dataset $D$ by leveraging an LLM agent $\mathcal{L}$ that alternates between two complementary phases: (1) \textsc{Action} ($\mathcal{L}_{\text{a}}$) and (2) \textsc{Reflection} ($\mathcal{L}_{\text{ref}}$).
In the \textsc{Action} phase, the agent generates an action conditioned on the task description, the current trajectory of steps taken, and the constitution (${OmniC}$) curated up to the previous \textsc{Reflection} phase.
In the \textsc{Reflection} phase, the agent produces reflections based on the task description, the observed trajectory, and the existing constitution.
These reflections include both task-agnostic long-term memories, shared across the environment, and task-specific progress updates, which act as short-term guidance for the current task.
Section~\ref{sec:ref_types} provides a detailed breakdown of these reflection types.

Actions are generated at each turn until either a terminal state is reached or a predefined maximum number of steps (${\text{turns}}_{\text{max}}$) is met. 
In contrast, reflections are generated at fixed intervals determined by a reflection frequency hyperparameter ($r_{\text{freq}}$), while long-term memory summarization is triggered based on a separate summarization frequency (${s}_{\text{freq}}$).
We adopt a hyperparameter-driven approach, rather than automating the invocation via LLM-as-a-Judge strategies \cite{zheng2023llmasjudge, gu2024surveyllmasjudge}, due to two main reasons: %\newline
% \begin{enumerate}
%     \item It avoids the inefficiencies of known pitfalls in LLM-as-a-judge based automation \cite{jain2024ragmodulo,gu2024surveyllmasjudge, li2025preferencebadllm,szymanski2025limitationsllmasjudge}
%     \item It offers greater control over the cost–performance trade-off of reflection.
% \end{enumerate}
(1) It avoids the inefficiencies of known pitfalls in LLM-as-a-judge based automation \cite{jain2024ragmodulo,gu2024surveyllmasjudge, li2025preferencebadllm,szymanski2025limitationsllmasjudge}; %\newline
  (2) It offers greater control over the cost–performance trade-off of reflection.%\newline

Specifically, agents might invoke reflection either excessively (seeking frequent reassurance) or insufficiently, due to misplaced confidence. 
Additionally, randomization of test episodes across benchmarks can skew task-specific insight distributions and complicate automated summarization timing, potentially causing under- or over-summarization.
We provide a detailed analysis of the effects of these hyper-parameters in Section~\ref{sec:ablation}.%\newline
% We analyze the effects of these hyper-parameters in detail in Section~\ref{sec:ablation}.\newline
% Section \ref{sec:ref_gen} further expands on details of reflection generation. 
\begin{algorithm}[H]%
    \caption{OmniReflect Methodology}
    \label{alg:omnireflect}
    \textbf{Input:}
    \begin{itemize}[noitemsep,topsep=0pt,parsep=0pt,partopsep=0pt]
        \item[\footnotesize{$\blacktriangleright$}] Dataset $D$ with set of tasks ${t_{i}}$, $D = \{{t_{i}} \mid$ ${i} \in \{1, \ldots, N\}$
        \item[\footnotesize{$\blacktriangleright$}] LLM Agent: $\mathcal{L}_a$, Reflection Agent: $\mathcal{L}_{\text{ref}}$
        \item[\footnotesize{$\blacktriangleright$}] Environment ${E}$ producing ($observation$, $reward$) on receiving $action$
    \end{itemize}        
    \textbf{Output:}
    \begin{itemize}[noitemsep,topsep=0pt,parsep=0pt,partopsep=0pt]
        \item[\footnotesize{$\blacktriangleright$}] Success Rate, ${SR}$
        \item[\footnotesize{$\blacktriangleright$}] Constitution, ${OmniC}$
        % \item[]
    \end{itemize}            
    \begin{algorithmic}[1]
      \STATE Initialize constitution, ${OmniC}$ %\gets [\ ]$  
      \STATE Intialize rewards, $rewards$ %\gets [\ ]$           
      \STATE Initialize hyper-parameters,\\ %\textbackslash
      \item [] $r_{\text{freq}}, s_{\text{freq}}, {\text{turns}}_{\text{max}}$
      %\item[] 
      \FOR{each task ${t_{i}}$ in $D$}
        \STATE Initialize trajectory, $\tau$% \gets [\ ]$
        \STATE Initialize task description, $d_{{t_{i}}}$ 
        \STATE Set $\text{turn} \gets 0$              
          \WHILE{($\text{turn} < {\text{turns}}_{\text{max}}$)}        
               \STATE $\triangleright$ \textit{Action Phase}
               \STATE $act \gets \mathcal{L}_a(d_{{t_{i}}}, OmniC, \tau)$
               \STATE $obs, reward \gets {E}(act)$
               \STATE Append $(act, obs)$ to $\tau$
               % \item[]
               \STATE $\triangleright$ \textit{Periodic Reflection Phase}
               \IF {($\text{turn}\ \%\ r_{\text{freq}}\ \text{is}\ 0$)}
                   \STATE $rules \gets  \mathcal{L}_{\text{ref}}(d_{{t_{i}}}, OmniC,\tau)$
                   \STATE Append $rules$ to $OmniC$
               \ENDIF
               \IF {$obs$ is final state}
                \STATE $\triangleright$  \textit{Task is complete}
                \STATE break
               \ENDIF
               \STATE Increment $\text{turn}$
          \ENDWHILE
          %\item[] 
          \STATE Append $reward$ to $rewards$
           \STATE $\triangleright$ \textit{Periodic Summarization}
           \IF {($i\ \%\ s_{\text{freq}}\ \text{is}\ 0$)}
               \STATE $OmniC \gets  \mathcal{L}_{\text{ref}}(OmniC)$
           \ENDIF                
      \ENDFOR
      %\item[] 
      \STATE $\triangleright$ \textit{Compute success rate}
      \STATE $SR \gets mean(rewards) * 100$          
      \RETURN $SR, OmniC$
    \end{algorithmic}
\end{algorithm}

\subsection{Reflection generation strategies}
\label{sec:ref_gen}
%In this work, we explore three distinct approaches to natural language reflection generation:
%
In this work, we explore three distinct approaches for generating natural language reflections based constitutions:
1) Neural 
2) Symbolic and 
3) Neuro-Symbolic.

\subsubsection{{Neural Generation}}
Following prior work \cite{shinn2024reflexion, madaan2024selfrefine}, we use LLMs to generate reflections.
We craft reflection-oriented prompts and use $\mathcal{L}_{\text{ref}}$ to produce a list of rules or insights, which collectively form the constitution.
%
% Each reflection step is conditioned on the current task context and any previously generated reflections, collected in their respective constitutions.
%
To avoid redundancy and ensure meaningful information gain, $\mathcal{L}_{\text{ref}}$ also periodically summarizes past reflections.
The constitution is continuously updated during task execution and leveraged in subsequent steps to guide the agent towards successful task completion. 

\subsubsection{{Symbolic Generation}}
We extend the base regular expressions introduced by AgentBoard \cite{ma2024agentboard} for progress tracking, enhancing them to support both fine-grained and high-level template-based natural language feedback.
Additionally, we design a concise set of task-specific and environment-level rules based on analysis of representative successful and failed trajectories.
% Furthermore, a small set of task-specific and envrionment level rules are devised based on analysis of samples of successful and unsuccessful trajectories.
% We extend base regular expressions created by AgentBoard \cite{ma2024agentboard} for progress tracking, to enable both fine-grained and high-level template-based natural language feedback. 
% Examples for all datasets are provided in Appendix \hl{TODO}.
% %
% Specifically, we extend base 
% Researchers are provided with a small calibration set containing both successful and failed trajectories, along with a few base regular expressions. 
% %
% They are then tasked with crafting additional expressions that enable both fine-grained and high-level reflections within OmniReflect. 
%
% Furthermore, they design system prompts, to emulate abstract constitutional rules that guide the behavior of the LLM agent at the environment level. 
%
During each reflection step, the current trajectory, particularly the most recent observation, is evaluated against these regular expressions to generate reflections.
% Illustrative examples for all datasets are provided in Appendix \hl{TODO}.

\subsubsection{{Neuro-Symbolic Generation}}
This method guides the Neural system using few-shot exemplars, which are derived from reflections produced during the Symbolic Generation process.
% This approach leverages symbolic artifacts from the \textit{Symbolic Generation} process as few-shot exemplars to guide the Neural system. 
%
This approach aligns LLM-generated reflections more closely with human intuition, while substantially reducing the need for extensive annotations typically required by reliable symbolic systems.
%
% In contrast to fully symbolic methods, we only require a single annotated set per task type rather than comprehensive coverage of the entire observation space.
%
% This hybrid strategy combines the interpretability of symbolic systems with the flexibility and scalability of neural models.

Refer to Appendix Section \ref{app:prompts} for prompt details and Section \ref{app:symbolic_reflections} for sample regular expressions used across benchmarks.

% \begin{figure}[h]
%     \centering
%     \includegraphics[width=0.9\linewidth]{images/omnireflect_constitution.png}
%     \caption{Overview of OmniReflect's three-tiered reflection strategy: (1) Environment-level Abstract reflections capture foundational generalizations; (2) Environment-level Error reflections enable early mistake detection and efficiency improvement; and (3) Task-level \textsc{progress} reflections provide fine-grained, task-specific updates that reinforce optimal task completion.}
%     \label{fig:reflection_types}
% \end{figure}

\subsection{Reflection categories}
\label{sec:ref_types}

The \textsc{Reflection} phase in OmniReflect produces three distinct types of reflections:  
(1) \textbf{Abstract},  
(2) \textbf{Error}, and  
(3) \textbf{Progress}.  
Figure~\ref{fig:omnireflect_overview} presents representative examples of each of these reflection types, drawn from sample tasks in the ALFWorld, BabyAI, and PDDL environments.

In the OmniReflect framework, Abstract and Error reflections are periodically summarized and stored in long-term memory, which is shared across tasks and sessions\footnote{While OmniReflect is initially designed for single-trial operation per task, it can be extented to multiple trials while preserving consistent knowledge sharing across trials.} at the environment level. 
In contrast, progress tracking reflections serve as transient, short-term knowledge that guides decision-making within the current task episode\footnote{For clarity and conciseness, three types of reflections are not mentioned within the algorithm.}.

\subsubsection{{Abstract Reflection}}
% We hypothesize that, when exposed to a novel environment with appropriate instructions, an LLM agent can effectively explore and document both the characteristics and limitations of that environment through natural language reflections.
%
We generate both \textit{task-specific} and \textit{task-agnostic} abstract reflections. 
In the \textit{task-specific} setting, the agent solves individual tasks by using reflection to align its existing knowledge with the unique aspects of the current environment.
This helps identify and correct errors caused by mismatched assumptions about the environment, often caused due to the implicit world knowledge of the model.
In the \textit{task-agnostic} setting, the agent{'}s objective is to explore the environment broadly to generate reusable environment-level knowledge that benefits future tasks. 
In the OmniReflect framework, \textit{task-specific} abstract reflections are generated in both the Self-sustaining and Co-operative modes. 
In contrast, \textit{task-agnostic} abstraction is utilized exclusively in the Co-operative mode to address the challenge of limited data availability.

\subsubsection{{Error Reflection}}
We build upon prior work such as Reflexion~\cite{shinn2024reflexion}, extending its strategy of trajectory analysis to generate actionable feedback for error correction.
Unlike existing methods that typically generate reflections post hoc or across multiple task trials \cite{renze2024self}, our OmniReflect framework introduces \textit{in-situ} reflections: guidance is generated at periodic intervals within a single task execution episode. 
This enables the agent to recover from mistakes more rapidly, reducing the number of iterations needed for successful task completion.
Furthermore, the agent evaluates the efficiency of its current trajectories, identifying recurring planning inefficiencies along with potential strategies for improvement. 

\subsubsection{{Progress Reflection}}
OmniReflect generates task-specific progress-tracking thoughts, to assist in monitoring the execution of necessary sub-tasks in an optimal order to complete tasks.
Unlike error reflections, they offer not only actionable guidance but also non-actionable, grounding observations that affirm completed sub-tasks, thereby reinforcing the agent{'}s confidence in progressing to subsequent steps.

\section{Experimental Details and Results}
We conducted experiments on ALFWorld \cite{shridhar2020alfworld}, BabyAI \cite{chevalier2018babyai}, and PDDL \cite{silver2020pddlgym}, three benchmark environments that balance navigation, reasoning, and compound task-solving.
Details of the publicly available datasets, model configurations, and experimental settings are provided in the following sections.
%
% We evaluate our approach across multiple benchmark environments, with experimental settings and dataset descriptions detailed in the following sections.
\subsection{Datasets}

% \noindent\textbf{ALFWorld} is a simulated home environment designed to evaluate the ability of an agent to perform household tasks through goal-directed navigation and interaction. 
\noindent\textbf{ALFWorld} is designed to evaluate the ability of an agent to perform household tasks through goal-directed navigation and interaction. 
We use the \textit{unseen split} of the dataset for our experiments. % to assess generalization. 
A detailed description is provided in Appendix~\ref{app:alfworld}.\newline

\noindent\textbf{BabyAI} is a grid-based environment in which agents navigate through interconnected minigrids to solve tasks such as “pick up a red box and then go through the grey door to the right”. 
We adopt the same test split used by AgentBoard \cite{ma2024agentboard}. 
See Appendix~\ref{app:babyai} for more details.\newline

\noindent\textbf{PDDL} contains planning benchmarks focused on task decomposition and state optimization. 
Our experiments include four domains: \textit{Gripper} (object transport between rooms), \textit{Blocksworld} (block stacking and unstacking), \textit{Barman} (cocktail preparation), and \textit{Tyreworld} (tool-based mechanical repairs). 
We follow the same test split as used by AgentBoard. 
Refer to Appendix~\ref{app:pddl} for details.

\subsection{LLMs and Experimental Setup}
\label{sec:exp_setup}

\noindent\textbf{Large Language Models.} Our study employed three widely recognized LLMs as agents: (1)  Qwen3-4B \cite{qwen3}, (2) Gemini-2.0 (including its “\textit{flash}" variant) \cite{google2025gemini} and (3) GPT-4 (including its “\textit{omni}" variant)  \cite{hurst2024gpt4o}.
% 
% Our study employed three LLMs agents: (1)  Qwen3-4B \cite{qwen3}, (2) Gemini-2.0 (including its “\textit{flash}" variant) and (3) GPT-4 (including its “\textit{omni}" variant)  \cite{hurst2024gpt4o}.
%
To ensure a balance between reproducibility and performance, we set the temperature to 0, the nucleus sampling probability ($top\_p$) to 0.7, the token sampling limit ($top\_k$) to 50, and applied a repetition penalty of 1.\newline

\noindent\textbf{ReAct.} The ReAct reasoning strategy \cite{yao2022react} combines reasoning (thinking) and acting within agentic environments, enabling step-by-step decomposition of complex tasks. 
For ALFWorld, we use the prompts provided by the original authors \cite{yao2022react}, while for BabyAI and PDDL, we adopt ReAct prompts crafted by AgentBoard. 
Each experiment uses a single trial with a maximum of 50 turns per task. \newline

\noindent\textbf{Reflexion.} While using Reflexion \cite{shinn2024reflexion} as our baseline, we adopt the same protocol as the original work, allowing up to 15 trials per task. 
 Each trial consists of 50 turns.\newline%, meaning that in the worst case, Reflexion may use up to \textit{764} LLM inferences per task. 
% This accounts for 50 steps per trial plus 14 reflection calls between trials. \newline\newline

\noindent\textbf{OmniReflect.}
We adapt the ReAct reasoning strategy with 1-2 few-shot examples to construct our base prompt. 
For the primary results reported in Table \ref{tab:omni_main}, we use a reflection frequency ($r_{freq}$) of 10 and a summarization interval ($s_{freq}$) of 10 as hyper-parameters. 
Specifically, we generate reflections and update the constitution every 10 turns, and perform a constitution summary after completing 10 tasks.
% The impact of these hyper-parameters is discussed in Section \ref{sec:ablation}. 
Similar to ReAct, OmniReflect uses a single trial with a maximum of 50 turns per task.
\begin{table*}
    \centering
    \resizebox{1\linewidth}{!}{
        \begin{tabular}{l||c|c|c||c|c|c||c|c|c}
        \hline
        % \multicolumn{10}{c}{\textbf{Success Rate (\%) }}\\\hline 
              &  \multicolumn{3}{c||}{\textbf{ALFWorld}}  &  \multicolumn{3}{c||}{\textbf{BabyAI}}  &  \multicolumn{3}{c}{\textbf{PDDL}} \\\hline
                            &  Qwen  & Gemini & GPT-4 &  Qwen  & Gemini & GPT-4 & Qwen  & Gemini & GPT-4 \\\hline
           ReAct   &  44.0              & 68.0               & 54.2              &  24.6              & 45.2              & 31.2             & 1.7              & 41.7               & {76.7} \\\hline
           Reflexion   &  82.8 & \textbf{94.0}   & 84.1  &  44.1  & 53.2   & 50.8  & 11.7  & 66.7  & \textbf{91.7} \\\hline \hline
           
           \multicolumn{1}{l}{\textbf{Self-sustaining mode}}  \\ \hline
           OmniReflect-Neural          &  83.6  & 91.8   & 94.8  &  \textbf{73.2}  & \textbf{74.1}   &  \textbf{72.3} & {20.0}  & 71.67  & 78.3\\
           OmniReflect-Symbolic       &  \textbf{91.8}  & 88.8   & \textbf{100.0}   &  45.5  & {67.9}   & 60.7 & 16.7  &  \textbf{78.3}  & \textbf{85.0}\\
           OmniReflect-Neuro-Symbolic      &  {86.6}  & \textbf{93.3}   & {96.3}  &  {54.5}  & 64.3   & {68.8} & \textbf{31.7}  &  {75.0}  & 80.0\\
           \hline \hline

           \multicolumn{1}{l}{\textbf{Co-operative mode with ReAct agent}} \\ \hline
           OmniReflect Meta-advisor\textsubscript{Qwen}         &  {73.1}  & {77.6}   & 93.8              & {58.0}   & {58.0}  & {59.8} & {12.1} & 36.6               & 70.0\\
           OmniReflect Meta-advisor\textsubscript{Gemini}       & 55.2                & 47.0              & {94.8}  & 50.9               & {58.0}   & 52.7             & 10.0            & {45.7}   &  75.0\\
           OmniReflect Meta-advisor\textsubscript{GPT-4}        &  \textbf{76.9}     & \textbf{79.1}      & \textbf{96.3}     &  \textbf{60.7}     & \textbf{63.4}     & \textbf{64.2}    &  \textbf{13.3}   & \textbf{65.2}      & \textbf{80.0} \\
           \hline

        \end{tabular}
    }
    \caption{Success Rate (\%) of different LLM-agents across ALFWorld, BabyAI, and PDDL environments. All results follow the experimental setup described in Section~\ref{sec:exp_setup}. Reflexion \cite{shinn2024reflexion} results indicate final performance after 15 trials are completed. All ReAct and OmniReflect results only use 1 trial. The highest-performing results are shown in bold. Qwen and Gemini refer to Qwen3-4B and Gemini-2.0 respectively.}
    \label{tab:omni_main}
\end{table*}
%--------------------------
\subsection{Self-sustaining mode: OmniReflect as an agent}
The primary results for the Self-sustaining mode are shown in Table~\ref{tab:omni_main}.
OmniReflect achieves the highest success rate in \textbf{7 out of 9} evaluated settings, outperforming Reflexion baselines under at least one of the Neural, Symbolic, or Neuro-Symbolic configurations. 
The only exceptions are Reflexion Gemini-2.0 on ALFWorld and Reflexion GPT-4 on PDDL. 
Averaged across models, OmniReflect yields substantial performance gains over Reflexion’s 15-trial setup, despite operating with only a single trial augmented by periodic reflection: +10.3\% on ALFWorld, +23.8\% on BabyAI, and +8.3\% on PDDL.

Performance patterns across environments further illuminate the strengths of each reflection strategy. 
On ALFWorld and PDDL, where tasks exhibit procedural regularities and structured action sequences, Symbolic and Neuro-Symbolic variants of OmniReflect consistently achieve top performance. 
These results highlight the strength of symbolic mechanisms, such as regular expressions, for progress monitoring and ensuring plan adherence. 
% Conversely, in BabyAI, where success is tightly coupled to dynamic agent states and spatial exploration (e.g., object and door placement is random), neural approaches dominate. 
Conversely, in BabyAI, where success is tightly coupled to dynamic spatial exploration (e.g., object and door placement is random), Neural approaches dominate. 
This suggests that flexible, open-ended neural reasoning is better suited for environments with high variability and partial observability.

Collectively, these results underscore the efficacy of hierarchical, in-session reflection during task execution, demonstrating its ability to enable early identification and resolution of errors, leading to significantly improved task completion across diverse environments.

\subsection{Co-operative mode: OmniReflect as a Meta-advisor}
\label{sec:meta-advisor}
% Table~\ref{tab:omni_main} reports success rate improvements for ReAct-based agents, both with and without an OmniReflect meta-advisor. 
%
Table~\ref{tab:omni_main} also reports success rate improvements achieved when using OmniReflect Meta-advisor models.
In this setup, the Meta-advisor constructs the constitution, while the consumer agent applies ReAct-style reasoning guided by the derived rules.
When equipped with constitutions distilled via OmniReflect from just one calibration example per task type, ReAct agents exhibit substantial performance gains, achieving average improvements of 28\% on ALFWorld, 29\% on BabyAI, and 20.9\% on PDDL, using GPT-4 as the Meta-advisor.
Crucially, these gains are realized without any additional LLM inferences at test time. Instead, the Meta-advisor generated constitutions are injected into the agent’s prompt, demonstrating that even a lightweight integration of constitutional guidance can yield strong downstream benefits. 
Notably, this setting utilizes only environment-level reflections (specifically Abstract and Error), without incorporating task-level progress tracking or dynamic reflection, as the ReAct agent does not performing on-the-fly reflective updates during task-execution.

The results underscore the effectiveness and transferability of OmniReflect-generated constitutions ($OmniC$), across different LLM backbones highlighting their scalability and versatility.

% The results underscore the effectiveness and transferability of OmniReflect-generated constitutions ($OmniC$), across different LLM backbones (including smaller models such as Qwen3-4B on PDDL) highlighting their scalability and broad applicability.
% The influence of the meta-advisor’s architecture is further explored in Section~\ref{sec:ablation}.

\subsection{Ablation Studies and Discussion}
\label{sec:ablation}
\noindent\textbf{Choice of OmniReflect Meta-advisor.}
% As illustrated in Table \ref{tab:omni_main}, the reasoning capabilities of a language model are strongly correlated with both its ability to generate high-quality constitutions and to follow them effectively. A larger and more capable model such as GPT-4 demonstrates exceptional performance as both a meta-advisor (Section~\ref{sec:meta-advisor}) and as a follower.
As illustrated in Table \ref{tab:omni_main}, reasoning capabilities are strongly correlated with both its ability to generate high-quality constitutions and to follow them effectively. A larger and more capable model such as GPT-4 demonstrates exceptional performance as both a Meta-advisor (Section~\ref{sec:meta-advisor}) and as a follower.
%, achieving the highest rates of constitution adherence and environment understanding when learning from smaller models like Qwen3-4B and Gemini-2.0.
Despite its smaller scale, Qwen3-4B proves to be a surprisingly competitive Meta-advisor, frequently enabling greater performance gains in downstream ReAct agents compared to Gemini-2.0.

In total, \textbf{24 out of 27} evaluated Meta-advisor/ReAct agent configurations show substantial performance improvements over the ReAct-only baseline, demonstrating the robustness and transferability of OmniReflect-derived constitutions across model architectures.
The only exceptions occur when Gemini-2.0 serves as both the ReAct agent and Meta-advisor on ALFWorld, and when GPT-4 ReAct agent is paired with either Qwen3-4B or Gemini-2.0 as Meta-advisors on PDDL.
These cases likely reflect either insufficient abstraction quality or weaker synergy in reflection transfer across model scales.

Notably, one compelling result is that a Qwen3-4B ReAct agent, when guided by GPT-4 as a Meta-advisor, outperforms all Reflexion baselines on BabyAI. 
This highlights that even smaller models can demonstrate advanced reasoning when guided by high-quality constitutions generated by a capable OmniReflect agent.\newline

\noindent\textbf{Impact of Reflection Hyper-Parameters.}
Table~\ref{tab:ablation_hpo} illustrates that across all datasets, over-summarization can degrade performance, potentially omitting useful information. 
In contrast, increasing the frequency of reflection generally yields greater benefits, particularly in partially observable environments like BabyAI, where ongoing self-analysis helps the agent better assess its progress and adapt rapidly.
% Only the OmniReflect-Neural setting is used for this experiment, due to its minimal reliance on human annotations, making it more readily adaptable. Furthermore, symbolic variants typically do not perform periodic reflection; instead, they trigger reflection conditionally based on the current state of progress.
This experiment uses only the OmniReflect-Neural setting, due to its minimal dependence on human annotations, making it more adaptable in practice. 
Additionally, the Symbolic variant typically do not perform periodic reflection; instead, they trigger reflection conditionally, based on the agent's current progress. \newline
\begin{table}
    \centering
    \resizebox{1\linewidth}{!}{
        \begin{tabular}{c|c||c||c}
        \hline
           \makecell{($r_{\text{freq}}, s_{\text{freq}})$} & \multicolumn{1}{c||}{\textbf{ALFWorld}} & \multicolumn{1}{c||}{\textbf{BabyAI}} & \multicolumn{1}{c}{\textbf{PDDL}}\\\hline
                                       (5, 5)   &  95.2 & 67.8  &  73.3 \\
                                       (5, 10)  &  96.7 & 65.2  & 76.7  \\
                                       (5, 20)  &  \textbf{97.0} & 65.2  &   78.3\\
                                       (10, 5) &  93.2 &  62.5 &  76.6\\       
                                       (10, 10) &  94.8 &  \textbf{72.3} & \textbf{78.3} \\       
                                       (10, 20) &  94.8 &  71.8 &  78.3\\
            \hline
        \end{tabular}
    }
    \caption{Success rates on ALFWorld, BabyAI, and PDDL using the OmniReflect-Neural setting with GPT-4, illustrating the impact of reflection and summarization hyperparameters on task performance.}
    \label{tab:ablation_hpo}
\end{table}

\noindent\textbf{OmniReflect Agent with OmniReflect Meta-advisor.}
Table \ref{tab:omniomni} demonstrates that both Neural and Neuro-Symbolic variants of the OmniReflect agent consistently outperform their ReAct counterparts, despite receiving identical guidance from the GPT-4 Meta-advisor to mitigate the cold-start challenge. 
This indicates that OmniReflect agents not only effectively integrate external advice but also adapt and refine it over time, exhibiting a robust capacity to evolve initial guidance into more performant strategies.\newline

\begin{table}
    \centering
    \resizebox{1\linewidth}{!}{
        \begin{tabular}{l||c||c||c}
        \hline
          & \multicolumn{1}{c||}{\textbf{ALFWorld}} & \multicolumn{1}{c||}{\textbf{BabyAI}} & \multicolumn{1}{c}{\textbf{PDDL}} \\\hline
 %& GPT-4 & GPT-4 & GPT-4 \\\hline
ReAct      & 96.3  &  64.2 &  80.0 \\
Neural     &  \textbf{100}    & \textbf{78.5}   &   86.7 \\
 Neuro-Symbolic        &   \textbf{100}   &  74.1   &  \textbf{90.0}  \\                 
            \hline
        \end{tabular}
    }
    \caption{Success rate~(\%) on ALFWorld, BabyAI, and PDDL using OmniReflect-Neural and OmniReflect-Neuro-Symbolic settings with GPT-4, highlighting the added benefit of leveraging pre-generated constitutions from a GPT-4-based Meta-advisor.}
    \label{tab:omniomni}
\end{table}

\begin{figure*}
    \centering
    \subfigure[ALFWorld]{\includegraphics[width=0.325\textwidth]{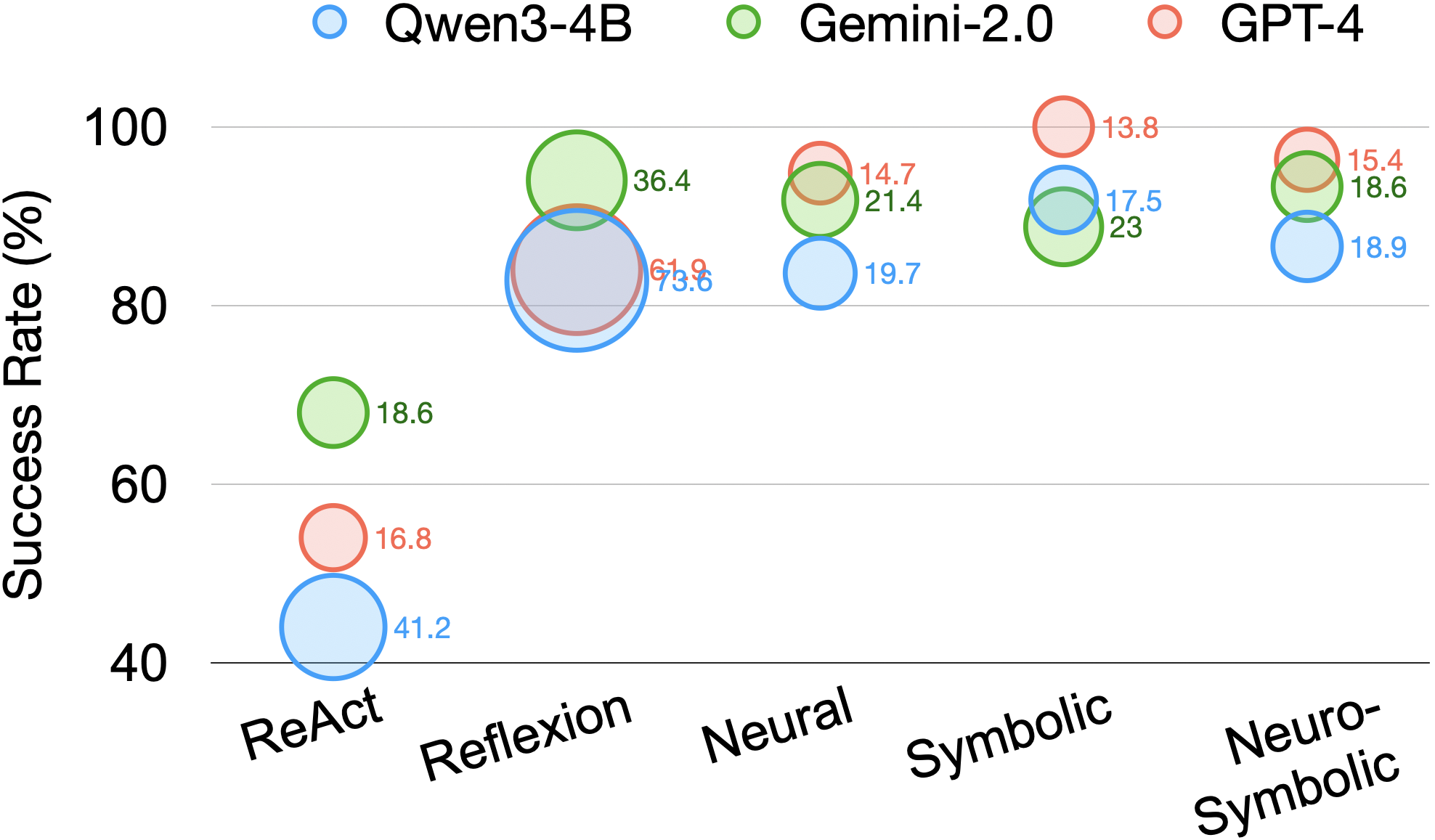}} 
    \subfigure[BabyAI]{\includegraphics[width=0.325\textwidth]{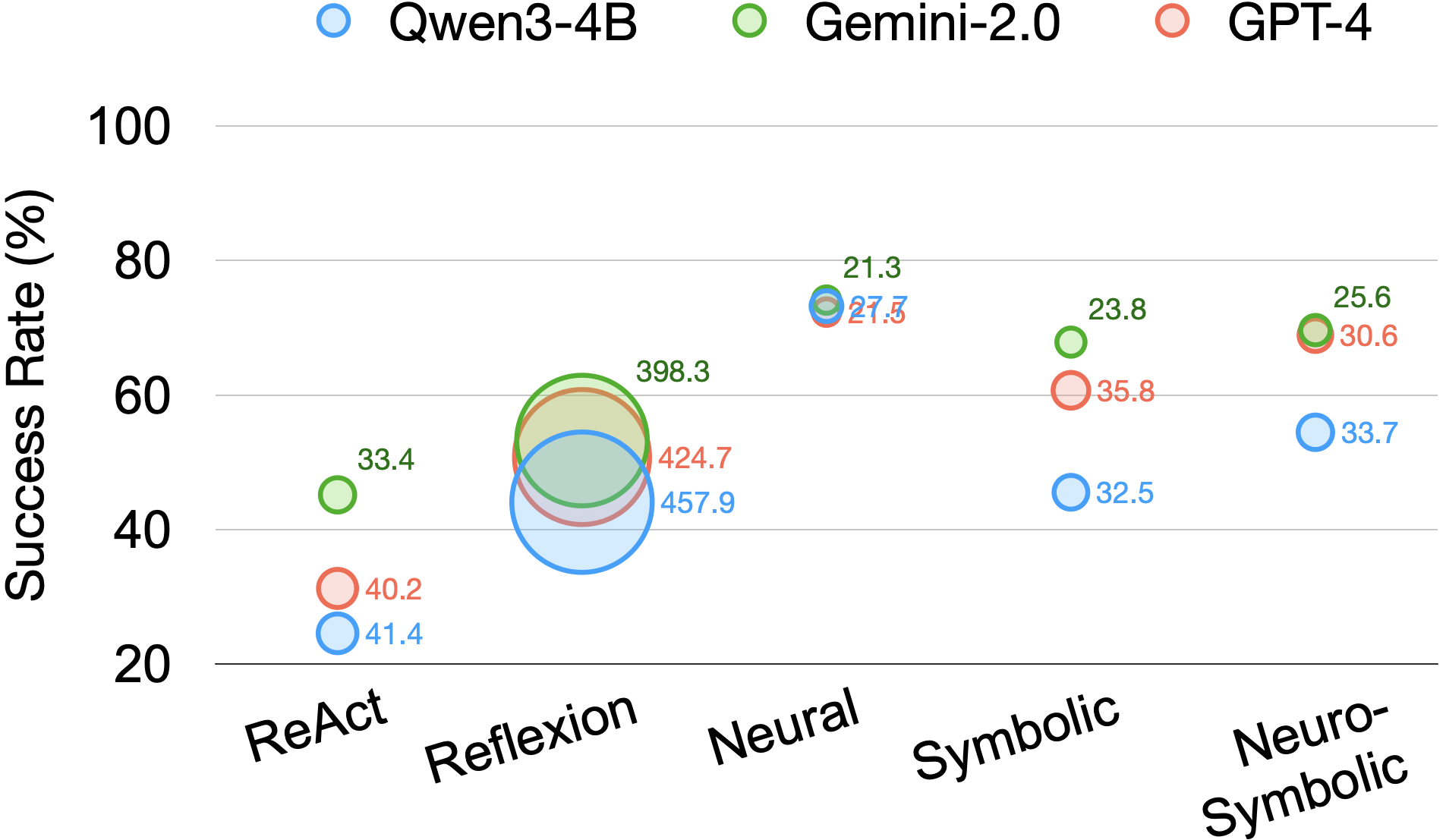}} 
    \subfigure[PDDL]{\includegraphics[width=0.325\textwidth]{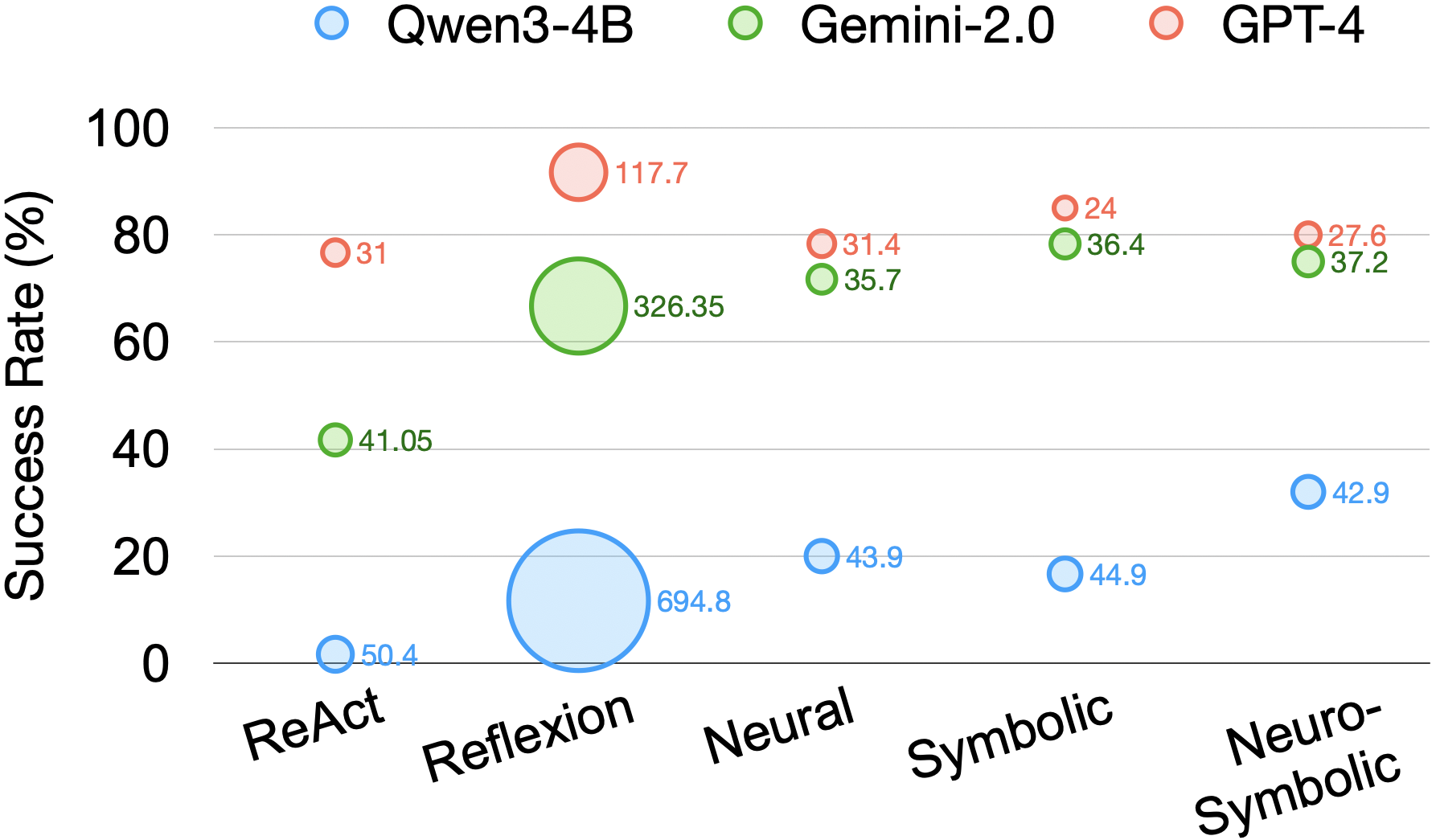}}
    \caption{Comparison of efficiency and effectiveness across ReAct, Reflexion, and OmniReflect agents reveals that OmniReflect achieves success rates comparable to or exceeding those of Reflexion, while maintaining inference efficiency on par with ReAct. Bubble sizes denote the average number of interaction turns per task.}
    \label{fig:call_efficiency}
\end{figure*}

\noindent\textbf{Cost Efficiency.}
Figure \ref{fig:call_efficiency} highlights the efficiency of the OmniReflect (Self-sustaining) approach compared to both the baseline Reflexion method and standard ReAct agents. 
% Across all datasets, OmniReflect consistently achieves performance comparable to or exceeding Reflexion, while maintaining inference efficiency on par with ReAct. 
The bubble sizes, reflecting the average number of interaction turns, demonstrate that OmniReflect agents not only match ReAct in operational efficiency, but also unlock effective self-correction mechanisms, consistently surpassing Reflexion in success rates across diverse tasks, datasets, and model scales.
% Specifically, the bubble sizes (representing the average number of interaction turns) show that OmniReflect agents, across all datasets, regardless of backbone size, operate with efficiency comparable to ReAct, while achieving substantially higher success rates than even Reflexion across a range of scenarios. 
Furthermore, OmniReflect's Neural and Neuro-Symbolic setting, when using multiple trials on ALFWorld (where Reflexion is the strongest) reach to 100\% in just one additional trial. 
These results underscore the inefficiencies introduced by task-specific, trial-level reflections and siloed knowledge\footnote{The average number of turns for Reflexion can be skewed due to a few failed examples. Nevertheless, the figure still provides an important comparison that OmniReflect is not impacted by outliers in the efficiency dimension.}.
In contrast, OmniReflect’s hierarchical reflection framework enables more generalizable and cost-effective learning, positioning it as a scalable and effective alternative for LLM-based agents.

Reflexion may requires up to 764 LLM calls per task, while our neural approach reduces this to under 80, achieving $\sim 700$ fewer inferences.
% Reflexion may require up to 764 LLM inferences per task, comprising 50 steps per trial and 14 reflection calls across trials.
% In contrast, our neural approach requires a base of 65 calls plus additional summarization calls (13 for ALFWorld, 11 for BabyAI, and 6 for PDDL), resulting in significantly ($\sim 700\downarrow$) fewer total LLM inferences. 
OmniReflect-Symbolic does not require any additional LLM calls to generate reflections. \newline
% \begin{figure*}
%     \centering
%     \subfigure[]{\includegraphics[width=0.45\textwidth]{images/reflexion_vs_us_sr_multi_trial.png}} 
%     \subfigure[]{\includegraphics[width=0.45\textwidth]{images/reflexion_vs_us_env_prop_multi_trial.png}} 
%     \caption{(a) Success rate over multiple trials (b) Proportion of enviromnets that are still to be solved at this trial - Placeholder until Nik's figure. Power of cold start shows the siginifcant difference - This could be appendix figure. just keeping as a reminder}
%     \label{fig:multitrials}
% \end{figure*}

\noindent \textbf{Impact of Types of Reflection.}
Table \ref{tab:ref_type_impact}, presents the contributions of the three distinct types of reflections employed in our study. 
Our analysis reveals that no single type of reflection emerges as the definitive leader, suggesting that their combined implementation is integral to the robust performance exhibited by our OmniReflect framework. 
Notably, the data underscores the critical role of environment-level Error reflection, particularly in scenarios where task-level Progress reflections are absent, i.e. the ReAct agent is integrated with the OmniReflect Meta-advisor. 
Owing to the intricate structure of PDDL, where an action is successful only when all necessary conditions are met (as illustrated in TyreWorld, where the precondition for loosening requires that the agent possesses a wrench, the nut on the hub is tight, and the hub is grounded), Abstract reflections play a pivotal role in explaining these nuances that are often challenging to discern solely through error analysis.

\begin{table}
    \centering
    \resizebox{1\linewidth}{!}{
        \begin{tabular}{l||c||c||c}
        \hline
          & \multicolumn{1}{c||}{\textbf{ALFWorld}} & \multicolumn{1}{c||}{\textbf{BabyAI}} & \multicolumn{1}{c}{\textbf{PDDL}} \\\hline
 %& GPT-4 & GPT-4 & GPT-4 \\\hline
OmniReflect-Neural     &  94.8 & 72.3  &  78.3 \\
(-) Abstract           &  91.8   & 48.2 & \textcolor{red}{71.6}  \\
(-) Error              &  92.5   & \textcolor{red}{46.4} &  76.7  \\                 
(-) Progress           &  \textcolor{red}{88.8}   & 54.5  &  75.0 \\   
            \hline
ReAct + MetaAdvisor    &  96.3  & 64.2  & 80.0  \\
(-) Abstract           &  92.5   & 61.6  &  \textcolor{red}{73.3} \\ 
(-) Error              &  \textcolor{red}{90.6}  & \textcolor{red}{52.7}  &   75.0 \\ 
\hline
        \end{tabular}
    }
    \caption{Success rate~(\%) on ALFWorld, BabyAI, and PDDL using OmniReflect-Neural and a ReAct agent with a Meta-advisor, illustrating the contributions of individual reflection types. The largest performance drops are highlighted in red. GPT-4 is used as both the agent and Meta-advisor in their respective settings.}
    \label{tab:ref_type_impact}
\end{table}

For an in-depth discussion on the influence of calibration data size, as well as analysis and illustrative examples of constitutions produced by various models across multiple experiments, please consult Appendix section \ref{app:results}.

% not enough space for subsection spacing
\section{Related Works}
\label{literature}
Constitutional AI \cite{bai2022constitutional} introduced the use of human-written constitutions to promote helpful and harmless behavior. In contrast, our framework autonomously curates task-oriented constitutions focused on improving robust task-completion quality. 
Moreover, unlike their finetuning-based approach, we leverage prompt-based guidance.

Self-correction methods like Self-Consistency \cite{wangself2023}, Universal Self-Consistency \cite{chen2024universal}, and MCR \cite{yoran2023answering} enhance reasoning by aggregating or meta-reasoning over multiple CoT paths.
Complementary work leverages iterative correction through natural language feedback \cite{madaan2024selfrefine, shinn2024reflexion}, numeric rewards and meta-feedback \cite{pan2024automatically}, and introspective learning via Self-Play Fine-Tuning in weaker LLMs \cite{chen2024self}.
Contrastively, OmniReflect performs reflection at both the environment and task level, using constitution-style rules.
It enables interpretable self‑improvement in a single trial (without multiple reasoning chains or repeated sampling) while markedly boosting weaker LLMs without extra fine‑tuning or inference overhead.

% Memory-based approaches like 
MemoryBank \cite{zhong2024memorybank}, RET-LLM \cite{modarressi2023ret}, and MemGPT \cite{packer2023memgpt} use structured memory or retrieval to persist knowledge, but face challenges like drift, size limits, and relevance filtering \cite{wu2024easily}. In contrast, OmniReflect maintains a compact, coherent memory via periodic constitution summarization, avoiding unbounded growth.

Automatic prompt construction approaches like \cite{shin2020autoprompt, zhang2022tempera, xu2022gps, prasad2022grips, li2021prefix, pryzant2023automatic, guo2309connecting, yang2023instoptima, tang2025unleashing} leverage LLMs as optimizers to adapt prompts for specific downstream tasks. 
Instead, our approach uses a straightforward strategy by appending constitutions to system prompts that guide the model in using them effectively.

\section{Conclusion}
%----------------------
% Smaller Version of conclusion 
%----------------------
We introduced OmniReflect, a hierarchical reflection-driven framework that summarizes task and environment-level insights into a reusable constitution, guiding LLM agents in complex environments.
%
% By alternating between action and reflection phases, and using periodic summarization, OmniReflect enables efficient long-term knowledge accumulation.
%
% While operating in the Self-sustaining mode, an OmniReflect agent independently orchestrates the creation and maintenance of its constitution while simultaneously accomplishing designated tasks. 
% Conversely, within the Co-operative mode, it formulates a constitution predicated on a calibration set, which subsequently can be used to boost the performance of another agent.
%
%
% It operates effectively through the Self-sustaining mode, wherein an autonomous agent independently synerzing constitution creation and maintenance along with task completion, and the Co-operative mode, in which a meta-advisor extrapolates constituion from a limited calibration dataset to direct the actions of another agent.
%
It operates effectively in Self-sustaining mode and Co-operative mode, where a constitution derived from minimal calibration significantly boosting smaller agent{'}s performance.
% It operates effectively in Self-sustaining and Co-operative modes, with constitutions derived from minimal calibration significantly boosting smaller agent{'}s performance.
%
Our Neural, Symbolic, and Neuro-Symbolic strategies balance adaptability with efficiency.
Empirical results across ALFWorld, BabyAI, and PDDL demonstrate consistent improvements over strong baselines, underscoring OmniReflect's scalability, generalizability, and cost-efficiency in enhancing self-reflection, and adaptability in LLM agents, serving as a crucial benchmark toward building more efficient and autonomous language-based agents.
% Empirical results across ALFWorld, BabyAI, and PDDL show consistent gains over strong baselines, highlighting OmniReflect{'}s scalability, generalizability, and cost-effectiveness in enhancing the reasoning, self-reflection, and adaptability of LLM agents, marking a step forward in building more efficient and autonomous language-based agents.
\clearpage
%----------------------
% Larger Version of conclusion 
%----------------------

% In this work, we introduced OmniReflect, a reasoning- and reflection-driven framework that leverages summarized reflections—termed constitutions—to guide LLM-based agents in solving complex tasks within dynamic environments.
% %
% By combining task-specific and environment-level reflections with periodic summarization, OmniReflect enables continual knowledge accumulation and reusability across sessions.
% %
% Our extensive empirical evaluations across challenging benchmarks (ALFWorld, BabyAI, and PDDL) demonstrate that OmniReflect consistently outperforms strong baselines while maintaining cost efficiency.
% %
% The framework operates effectively in both Self-sustaining and Co-operative modes, and we find that constitutions generated from a small calibration set can significantly enhance the performance of even smaller LLMs, underscoring OmniReflect’s versatility as a constitution generator.
% %
% To support efficient reflection generation, we explore neural, symbolic, and neuro-symbolic strategies—striking a balance between adaptability and interpretability.
% %
% Overall, OmniReflect establishes a scalable and generalizable paradigm for improving the reasoning, self-reflection, and adaptability of LLM agents, marking a step forward in building more efficient and autonomous language-based agents.

\section{Limitations}
While OmniReflect delivers strong performance gains, it introduces additional LLM calls, which may pose challenges for real-world deployment. 
However, we show that constitutions generated by smaller models (e.g., Qwen3-4B) can significantly improve the performance of larger models like GPT-4 and Gemini-2.0, suggesting that overhead can be mitigated through strategic model selection. 
Currently, constitutions are integrated without filtering, which may increase computational costs for models with limited context windows and introduce noise. 
Future work will explore more efficient constitution integration to reduce overhead and enhance usability. 
Furthermore, multi-constitution integration offers a promising direction for leveraging diverse pre-training strategies across multiple foundation models, resulting in greater variety in the generated constitutions.
Though we evaluate OmniReflect in embodied agentic settings, extending it to broader reasoning and planning tasks remains a promising direction. 
Finally, while we use ReAct for its simplicity and minimal inference overhead, future efforts will explore combining OmniReflect with advanced strategies such as Self-Consistency to further strengthen agent robustness.
\bibliography{custom}
\clearpage
\appendix
\label{sec:appendix}
% \subfile{appendix_arxiv}
\section{Additional Results and Discussion}
\label{app:results}
\subsection{Amount of calibration data.}
Table~\ref{tab:ablation_cal_data} shows that increasing the amount of calibration data generally improves the quality of the meta-advisor, thereby enhancing the downstream performance of ReAct agents that rely on it. However, performance gains begin to taper beyond a certain point.
We exclude BabyAI and PDDL from this analysis, as using more than one example per task type would constitute nearly 50\% of their respective test sets, undermining the goal of demonstrating that the meta-advisor can be calibrated with significantly fewer examples than required for evaluation. A calibration factor beyond 5 approaches this threshold for ALFWorld as well, and thus serves as the upper bound in our experiments.

% Figure \ref{fig:app_cal_factor} shows that more abstract reflections are generated in proportion to the calibration factor, that is, the number of examples used per task type. However, this gain saturates as the number of examples increases from 3 to 5.

\begin{table}[H]
    \centering
    \resizebox{1\linewidth}{!}{
        \begin{tabular}{c|c||c|c}
        \hline
              \multicolumn{4}{c}{\textbf{ALFWorld - Success Rate}} \\\hline
            \makecell{Calibration \\Factor} & \makecell{Calibration Set\\ Size} &  Qwen  & GPT-4 \\\hline
           1      &   6     & 76.9 & 96.3\\
           3      &   18    & 79.1 & 97.8\\
           5      &   30    & \textbf{80.6} & \textbf{98.5} \\\hline
        \end{tabular}
    }
    \caption{Effect of the calibration factor (number of examples per task type) on constitution quality and downstream ReAct agent performance, as measured on the ALFWorld dataset. GPT-4 is used as the meta-advisor.}
    \label{tab:ablation_cal_data}
\end{table}

% \begin{figure}[th]
%     \centering
%     \includegraphics[width=1\linewidth]{images/abstract_alfworld_cal_size.pdf}
%     \caption{Calibration factor trends for Abstract reflections across task types in ALFWorld, with GPT-4 serving as the meta-advisor.}
%     \label{fig:app_cal_factor}
% \end{figure}

\subsection{Constitutions}
This subsection presents representative examples spanning all three datasets and reflection types. On average, all models generate the highest number of Abstract reflections (20–50 per dataset), while Error reflections at the environment level are less frequent (typically fewer than 20). In contrast, the number of Progress reflections scales with task complexity, as shown in the tables. GPT-4 consistently produced well-structured outputs, whereas Gemini 2.0 and Qwen3-4B encountered JSON-style formatting issues in over 50\% of cases, necessitating complex post-processing to recover structured data.

An exception was observed with Gemini 2.0, which generated nearly a hundred Error reflections for ALFWorld, diluting the effectiveness of targeted reflection and potentially contributing to its lower performance when guided by its own constitutions. GPT-4 produced the largest constitutions, often exhibiting high verbosity. While Gemini 2.0 and Qwen3-4B generated a comparable number of reflections, Qwen3-4B frequently yielded more coherent and concise summaries without sacrificing quality.

Notably, (without explicit guidance) Gemini 2.0 included priority annotations in its rules—for example:
{`priority': 2, `rule': `Prioritize checking locations where target objects are most likely to be found (e.g., drawers, shelves, cabinets, countertop).'}, indicating an attempt to encode further structure within its reflective outputs that can be leveraged for reasoning.

Table \ref{tab:alfworld_cnst}, Table \ref{tab:babyai_cnst}, and Table \ref{tab:pddl_cnst}, show examples of different types of constitutions created by all three models on ALFWorld, BabyAI, and PDDL respectively. 
We have used majority voting to choose constitution samples.
Model-specific reflection examples, when shown, are annotated to indicate their source.

\begin{table*}
\lstset{
    backgroundcolor=\color[RGB]{255,255,255},
    breaklines=true,
    breakindent=0pt,
    basicstyle=\ttfamily\small,
    frame=trbl,
    numbers=none,
    frameround = tttt
}\begin{lstlisting}
Abstract:
Use fridge for cooling
...
heat [object] with microwave [location] requires microwave to be closed
...
Plates can be found on countertops (Task Agnostic)
...
Put [object] in/on [location] if [object] is in inventory and [location] is accessible and [location] is reachable
...

Error:
** GPT-4o Example
{
    "mistake": "Went to locations that are not present in the environment.",
    "solution": "Carefully check the available locations before moving"
} ...
** Gemini Example
{
  "rule": "Close containers (e.g., fridge, microwave, cabinet, drawer) after use.",
  "priority": 7
} ...
** Qwen Example
{
   "check_locations_in_order_of_likelihood": "check locations in order of likelihood to improve efficiency."
} ...

...
Progress:
[
You have located an apple,
...
You have reached the microwave,
...
]
\end{lstlisting}
    \caption{ALFWorld Constitution Examples}
    \label{tab:alfworld_cnst}
\end{table*}

\begin{table*}
\lstset{
    backgroundcolor=\color[RGB]{255,255,255},
    breaklines=true,
    breakindent=0pt,
    basicstyle=\ttfamily\small,
    frame=trbl,
    numbers=none,
    frameround = tttt
}\begin{lstlisting}
Abstract:
If you encounter a barrier while moving forward, turn left or right to explore a different direction.
...
If you encounter a closed door, use the 'toggle and go through' command to open it and proceed.
...
If you see multiple doors, prioritize the closest one first.
...
If you see an object, note its color and position for future reference.

Error:
** GPT-4o Example
{
    "mistake": "Attempted to move forward into a barrier",
    "solution": "Should have turned right first to explore the room further"
}
...
{
    "mistake": "Attempted to open the door with an unrecognized action",
    "solution": "Should have checked valid actions before attempting to open the door"
} ...
** Gemini Example
{
  'mistake': 'The agent moved forward repeatedly without finding the red ball, even when facing a wall. This indicates inefficient exploration.',
  'solution': 'After hitting a barrier, the agent should turn left or right to explore other directions. The agent should also prioritize finding the red ball and use the `go to red ball` action if available.'
}...
** Qwen Example
{
  'suggestion': 'always check available actions before executing any movement or interaction command to ensure the action is valid'
}...

Progress:
[
You have found a blue key, now find a blue door.
...
]

\end{lstlisting}
    \caption{BabyAI Constitution Examples}
    \label{tab:babyai_cnst}
\end{table*}

\begin{table*}
\lstset{
    backgroundcolor=\color[RGB]{255,255,255},
    breaklines=true,
    breakindent=0pt,
    basicstyle=\ttfamily\small,
    frame=trbl,
    numbers=none,
    frameround = tttt
}\begin{lstlisting}
Abstract:

Gripper Example
If both grippers are occupied, move to the target room to drop the objects.
...
Blockworld Example
If the robot arm is holding a block, it can put down the block or stack it on another clear block.
...
Barman Example
If you need to transfer an ingredient from a shot glass to a shaker, ensure the shaker is clean and at the appropriate level.
...
Tyreworld Example
Complete the process on one hub before moving to the next, including jacking down the hub after replacing the wheel and tightening the nuts.

Error:

{
        "mistake": "Attempted to shake a cocktail without all ingredients in the shaker",
        "solution": "Ensure all required ingredients are in the shaker before shaking"
}
      ...
{
        "mistake": "Inefficient sequence of actions",
        "solution": "Plan the sequence of actions to minimize the number of steps, such as filling all ingredients in the shot glass before transferring to the shaker"
}

Progress:

[
  "You have moved to roomb with ball1 and ball2, now you should drop ball1 and ball2 in roomb.",
  "After dropping ball1 and ball2, you should move back to rooma to pick up ball3 and ball4.",
  "Once you have picked up ball3 and ball4, move to roomb and drop them there.",
  "After dropping ball3 and ball4, return to rooma to pick up ball5 and ball6.",
  "Finally, move to roomb and drop ball5 and ball6 to complete the task."
]
...
[
  "You have been repeatedly attempting to unstack b5 from b3, which is not a valid action. Instead, consider other valid actions.",
  "Since b5 is clear and the robot arm is empty, you should pick up b5.",
  "After picking up b5, you can put it down on the table to free up b3.",
  "Once b3 is clear, you can unstack b3 from b4.",
  "After unstacking b3 from b4, you can put b3 on the table to free up b4.",
  "Then, you can unstack b4 from b2.",
  "After unstacking b4 from b2, you can put b4 on the table to free up b2.",
  "Next, you can unstack b2 from b1.",
  "After unstacking b2 from b1, you can put b2 on the table to free up b1.",
  "Now, you can stack b1 on b2 to satisfy the first goal condition.",
  "Then, you can pick up b4 and stack it on b1 to satisfy the third goal condition.",
  "Finally, you can pick up b3 and stack it on b5 to satisfy the second goal condition."
]
...

\end{lstlisting}
    \caption{PDDL Constitution Examples}
    \label{tab:pddl_cnst}
\end{table*}

\section{ALFWorld}
\label{app:alfworld}
\label{sec:appendix}

This section provides additional details and experimental results for the sequential decision making dataset ALFWorld. 
The embodied tasks are categorized into six types: Pick, Examine, Heat, Cool, Clean, and Pick Two. These tasks involve navigating a home environment to achieve specific goals, such as “\textit{place the vase in the safe}” or “\textit{inspect the book under the desk lamp}.”
Appendix Table \ref{app:ex:alfworld_put_task} provide a randomly chosen example annotation for a Pick task, along with a trajectory that solves the task. 

\begin{table*}[h]
\lstset{
    backgroundcolor=\color[RGB]{255,255,255},
    breaklines=true,
    breakindent=0pt,
    basicstyle=\ttfamily\small,
    frame=trbl,
    numbers=none,
    frameround = tttt
}\begin{lstlisting}
Task: Find some apple and put it in sidetable.
Initial observation: You are in the middle of a room. Looking quickly around you, you see a cabinet 4, a cabinet 3, a cabinet 2, a cabinet 1, a coffeemachine 1, a countertop 1, a diningtable 3, a diningtable 2, a diningtable 1, a drawer 1, a fridge 1, a garbagecan 1, a microwave 1, a sidetable 1, a sinkbasin 1, a stoveburner 4, a stoveburner 3, a stoveburner 2, a stoveburner 1, and a toaster 1. \newline Your task is to: find some apple and put it in sidetable.
Action 1: think: To solve the task, I need to find and take an apple, then put it in sidetable.
Observation 1: OK.
Action 2: think: First I need to find an apple. An apple is more likely to appear in fridges (1), diningtables (1-3), sidetables (1), countertops (1), sinkbasins (1), garbagecan (1). I can check one by one, starting with fridge 1.
Observation 2: OK.
Action 3: go to  fridge 1.
Observation 3: The fridge 1 is closed.
Action 4: open fridge 1.
Observation 4: You open the fridge 1. The fridge 1 is open. In it, you see a lettuce 2, a mug 2, a potato 2, and a tomato 1.
Exploring: ...
Action 11: go to  garbagecan 1.
Observation 11: On the garbagecan 1, you see a apple 3, and a egg 3.
Action 12: think: Now I find an apple (3). Next, I need to take it.
Observation 12: OK.
Action 13: take apple 3 from garbagecan 1.
Observation 13: You pick up the apple 3 from the garbagecan 1.
Action 14: think: Now I find an apple (3). Next, I need to put it in/on sidetable 1.
Observation 14: OK.
Action 15: go to  sidetable 1.
Observation 15: On the sidetable 1, you see a cup 1, a lettuce 1, a peppershaker 3, a potato 1, and a saltshaker 1.
Action 16: put apple 3 in/on sidetable 1.
Observation 16: You put the apple 3 in/on the sidetable 1.
\end{lstlisting}
\caption{ALFWorld: PUT Task}\label{app:ex:alfworld_put_task}
\end{table*}

\section{BabyAI}
\label{app:babyai}
\begin{figure}[H]
    \centering\resizebox{1.0\linewidth}{!}{
    \includegraphics[scale=0.6]{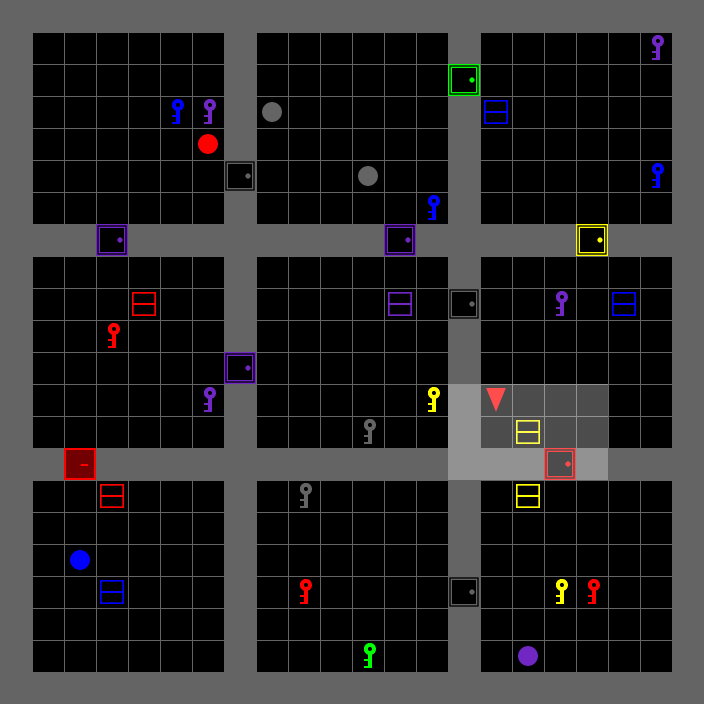} 
    }
    \caption{Visualization of a BabyAI grid environment showcasing balls, boxes, keys, and doors, with the red triangle marking the agent's location and orientation.}
    \label{fig:babyai_example}
\end{figure}

BabyAI environment was introduced in \cite{chevalier2018babyai} and covers tasks to be performed in a grid environment.
They can have multiple grid and minigrid sizes, ranging from a single minigrid to upto 9 minigrids. 
The minigrids can be of sizes 4$*$4 to 7$*$7. 
In this environment, the agent can see a 7$*$7 grid in the direction it is currently facing.
In most of the experiments, the agent is only exposed to this information which severely limits the global perspective of the complete grid which leads to lot of blind exploration.
Figure \ref{fig:babyai_example} provides an example of grid structure and objects used in BabyAI environment.

\section{PDDL}
\label{app:pddl}
PDDL benchmark was made accesible using \cite{silver2020pddlgym}. It contains four distinct environments that are used for this work: Gripper, Blockworld, Barman, and Tyreworld. 
Gripper and Blockworld provide an initial state and a goal state without explicit task instructions. 
However Barman and Tyreworld provide explicit task goals.
The agent is expected to reason, plan, and navigate to achieve the goal state. 
Examples of each dataset can be found in the Table \ref{tab:app:pddl_example} below. 

\begin{table*}
    \centering
\resizebox{1.0\linewidth}{!}{    \begin{tabular}{c|l}
    \hline
    Dataset & Example Task \\\hline
        Gripper & The goal is to satisfy the following conditions: ball1 is at roomb. ball2 is at roomb. ball3 is at roomb. ball4 is at roomb. \\
        Blockworld & The goal is to satisfy the following conditions: b1 is on b2. b2 is on b6. b3 is on b7. b5 is on b3. b6 is on b5. b7 is on b4.\\
        Barman & The goal is to satisfy the following conditions: shot1 contains cocktail2. shot2 contains cocktail3.\\
        Tyreworld & The goal is to satisfy the following conditions: Wheel r1 is inflated. r2 is on the-hub2. w1 is in boot. \\\hline
    \end{tabular}}
    \caption{Examples of sample tasks from datasets comprised in PDDL}
    \label{tab:app:pddl_example}
\end{table*}

\section{Prompts}
\label{app:prompts}
All prompts used in our experiments are outlined in this subsection.
\subsection{System Prompts}
System prompts are specific to the environment which outline action templates, and high level guidelines for solving tasks in the environment. 
Table \ref{app:ex:alfworld_cot_prompt} shows an example for ALFWorld, Table \ref{app:ex:babyai_cot_prompt} shares an example for BabyAI, and finally Table \ref{app:ex:pddl_cot_prompt} presents system prompt for PDDL.

\begin{table*}
\lstset{
    backgroundcolor=\color[RGB]{255,255,255},
    breaklines=true,
    breakindent=0pt,
    basicstyle=\ttfamily\small,
    frame=trbl,
    numbers=none,
    frameround = tttt
}\begin{lstlisting}
Interact with a household to solve a task.
You need to generate actions that strictly follow the below templates:
1. goto [location] 
2. take [object] from [location] put [object] in/on [location]
3. open [something]
4. close [something]
5. toggle [object][location]
6. clean [object] with [something]
7. heat [object] with [receptacle]
8. cool [object] with [receptacle]

Here are some aspects you have learnt so far.
** Abstract Reflections ** 

Here are some mistakes you have done so far, and potential solutions that can be used in next turns. 
** Error Reflections **

Here are some feedback about your progress so far and guidelines for next steps.
** Progress Reflections **

Here are two examples. They are very relevant. Please use the actions in these examples as your guidelines. 
\textit{Example 1: Truncated}
\textit{Example 2: Truncated}
\end{lstlisting}
\caption{ALFWorld Our Prompt}
\label{app:ex:alfworld_cot_prompt}
\end{table*}

\begin{table*}
\lstset{
    backgroundcolor=\color[RGB]{255,255,255},
    breaklines=true,
    breakindent=0pt,
    basicstyle=\ttfamily\small,
    frame=trbl,
    numbers=none,
    frameround = tttt
}\begin{lstlisting}
You are placed in a room and you need to accomplish the given goal with actions. 

You can use the following actions: 

- turn right 
- turn left 
- move forward 
- go to <obj> <id> 
- pick up <obj> <id> 
- go through <door> <id>: <door> must be an open door. 
- toggle and go through <door> <id>: <door> can be a closed door or a locked door. If you want to open a locked door, you need to carry a key that is of the same color as the locked door. 
- toggle: there is a closed or locked door right in front of you and you can toggle it. 

Here are some aspects you have learnt so far.
** Abstract Reflections ** 

Here are some mistakes you have done so far, and potential solutions that can be used in next turns. 
** Error Reflections **

Here are some feedback about your progress so far and guidelines for next steps.
** Progress Reflections **

Here are two examples. They are very relevant. Please use the actions in these examples as your guidelines. 
\textit{Example 1: Truncated}
\textit{Example 2: Truncated}
\end{lstlisting}
\caption{BabyAI Our Prompt}
\label{app:ex:babyai_cot_prompt}
\end{table*}

\begin{table*}
\lstset{
    backgroundcolor=\color[RGB]{255,255,255},
    breaklines=true,
    breakindent=0pt,
    basicstyle=\ttfamily\small,
    frame=trbl,
    numbers=none,
    frameround = tttt
}\begin{lstlisting}
---------blockworld---------
The robot has four actions: pickup, putdown, stack, and unstack. The domain assumes a world where there are a set of blocks that can be stacked on top of each other, an arm that can hold one block at a time, and a table where blocks can be placed.
    The actions defined in this domain include:
    pickup <block>: pick up a clear block
    putdown <block>: put down a block on the table 
    stack <block> <block>: stack a block on top of another block.
    unstack <block> <block>: unstack a block from on top of another block
---------barman---------
You are a robot barman that manipulates drink dispensers, shot glasses and a shaker. You have two hands. The goal is to find a plan that serves a desired set of drinks. Here are the actions you can do. Each valid action is a short phrase following fixed patterns:

    <hand> grasp <container>: Grasp a container
    <hand> leave <container>: Leave a container on the table
    fill-shot <shot> <ingredient> <hand1> <hand2> <dispenser>: Fill a shot glass with an ingredient from dispenser
    refill-shot <shot> <ingredient> <hand1> <hand2> <dispenser>: Refill a shot glass with an ingredient from dispenser
    empty-shot <hand> <shot> <beverage>: Empty a shot glass
    ...
    shake <cocktail> <ingredient1> <ingredient2> <shaker> <hand1> <hand2>: Shake a cocktail in a shaker
    pour-shaker-to-shot <beverage> <shot> <hand> <shaker> <level1> <level2>: Pour a beverage from a shaker to a shot glass from level1 to level2
---------gripper---------
You are a robot with a gripper that can move objects between different rooms. Your name is Robby.
    There are three actions defined in this domain:
    move <room1> <room2>: This action allows the robot to move from one room to another.
    pick <obj> <room> <gripper>: This action allows the robot to pick up an object using the gripper.
    drop <obj> <room> <gripper>: This action allows the robot to drop an object that it is carrying.
---------tyreworld---------
Your goal is to replace flat tyres with intact tyres on the hubs. Remember to open boot first to get tools you need. Intact tyres should be inflated. The nuts should be tight on the hubs. The flat tyres, wrench, jack, and pump should be in the boot. The boot should be closed.
    There are 13 actions defined in this domain:
    open <container>
    close <container>
    fetch <object> <container>
    put-away <object> <container>
    tighten <nut> <hub>
    jack-up <hub>
    jack-down <hub>
    undo <nut> <hub>
    do-up <nut> <hub>
    remove-wheel <wheel> <hub>
    put-on-wheel <wheel> <hub>
    inflate <wheel>


Here are some aspects you have learnt so far.
** Abstract Reflections ** 

Here are some mistakes you have done so far, and potential solutions that can be used in next turns. 
** Error Reflections **

Here are some feedback about your progress so far and guidelines for next steps.
** Progress Reflections **

Here are two examples. They are very relevant. Please use the actions in these examples as your guidelines. 
\textit{Example 1: Truncated}
\textit{Example 2: Truncated}
\end{lstlisting}
\caption{PDDL Our Prompt}
\label{app:ex:pddl_cot_prompt}
\end{table*}

\subsection{Symbolic Prompts}
Table \ref{app:tab:alfworld_symbolic}, Table \ref{app:tab:babyai_symbolic}, Table \ref{app:tab:pddl_symbolic} provide symbolic system prompts used by our OmniReflect-Symbolic System to provide environment level guidance.

\begin{table*}
    \centering
    {
\lstset{
    backgroundcolor=\color[RGB]{255,255,255},
    breaklines=true,
    breakindent=0pt,
    basicstyle=\ttfamily\small,
    frame=trbl,
    numbers=none,
    frameround = tttt
}\begin{lstlisting}
You need to go to a location or an object before using it or placing the objects at the location.
For example you need to `go to garbagecan 1` or `go to microwave 1` before using or placing the objects at the `garbagecan 1` or `microwave 1

You can only pick up or hold one object at a time. 

Everything in the environment is labelled with a numbers. You ALWAYS need to use the number that follows when referring to anything in the environment.
Valid example:`take lettuce 1 from countertop 1`
Invalid example: `take lettuce from countertop 1

You MUST Alternate between Thinking and Action generation. An example of think is `think: CD can be found on desk.` and An example of action is `take cd 1 from desk 1.`

You can ONLY use microwaves for heating. Once you are at a microwave, you can directly try to heat the item.
For example: For the action `go to microwave 1' can directly be followed by the action `heat apple 1 with microwave 1'

Once you are at a fridge, you can directly try to cool the item.
For example: For the action `go to fridge 1' can directly be followed by the action `cool lettuce 1 with fridge 1'

For tasks involving look or examine using desklamp you need to find a desklamp.
Once you are at a location with desklamp you can directly use the desklamp. The correct usage is through action of `use desklamp 1' for using desklamp 1.

For clean or cleaning tasks first obtain the item to be cleaned. You need to then clean the item at sinkbasin.
Once you are at a sinkbasin, you can directly try to clean the item.
For example: For the action `go to sinkbasin 1' can directly be followed by the action `clean plate 1 with sinkbasin 1'
\end{lstlisting}
}  
\caption{ALFWorld Symbolic System Prompt}
\label{app:tab:alfworld_symbolic}
\end{table*}

\begin{table*}[h]
\lstset{
    backgroundcolor=\color[RGB]{255,255,255},
    breaklines=true,
    breakindent=0pt,
    basicstyle=\ttfamily\small,
    frame=trbl,
    numbers=none,
    frameround = tttt
}\begin{lstlisting}
You are in a grid environment with multiple multiple minigrids.
Doors connect different mini grids that are separated by walls. You should go through doors if necessary to get to the destination.
For instance, if you are at row 4 and column 3 , facing up, and your target is at row 15 column 1, you should find a path to go down to row 15 and left to column 1 by toggling doors in between as needed.
YOU NEED KEY ONLY IF THE DOOR IS LOCKED. If a door is locked then you should find the same colored key to unlock and go through the door.
You only need a key once to toggle through the door. In the next turns, the door is no longer locked, do you do not need to pick up that color keys unnecessarily.

You can ONLY hold only one object at a time. If you are able to pickup an object, then drop what you are currently holding and then pickup the new object.

If you are facing a wall, turn left or turn right to explore other objects
If you to navigate to an object behind you, you can turn back. For example, If you are in minigrid 0, with direction ^ then turn back to access rest of the grid.

DO NOT repeat the turn multiple times, because you will get lost.

If you are blocked or having trouble picking up an object, you MUST turn to an empty cell and drop what you are currently holding, you CANNOT drop at the same location, as you are facing an object and then pickup the blocking object and move it out of the way. 
Once path is clear, you move or try to pick up the object that is blocking you.
You can ONLY drop objects in empty spaces. DO NOT DROP keys before you use them on the same colored door. You should drop them after toggling through the door
You MUST NOT drop an object immediately, as that would mean you are dropping it in the same place. So you MUST turn to an empty spot and then drop it. DO NOT DROP it in a cell that blocks your path to the next step.

-----------------------

First Turn: You should first generate a thought with a path from your minigrid to the destination minigrid with all the doors you need to go through. First determine, which door you should use to exit your grid if needed. For example, To go from minigrid 0 to minigrid 5, I need to go through yellow closed door 1 .....\n Generate this in less than 6 lines.

\end{lstlisting}
\caption{BabyAI Symbolic System Prompt}
\label{app:tab:babyai_symbolic}
\end{table*}

\begin{table*}[h]
\lstset{
    backgroundcolor=\color[RGB]{255,255,255},
    breaklines=true,
    breakindent=0pt,
    basicstyle=\ttfamily\small,
    frame=trbl,
    numbers=none,
    frameround = tttt
}\begin{lstlisting}
These are just guidelines and not the complete commands, so you should generate a correct command in the correct template.
If your subgoal is that a shot contains an ingredient, you should do the following steps:
    1. grasp the correct shot
    2. fill-shot using the dispenser that contains the ingredient

If your subgoal is that a shot contains a cocktail, you should do the following steps:
    Phase 1: Collecting all ingredients into a shaker, for each ingredient in the cocktail do the following
        1. grasp the correct shot
        2. fill-shot using the dispenser that contains the ingredient
        3. pour-shot-to-clean-shaker
        4. clean-shot
    Phase 2: Shake and serve
        1. leave the shot
        2. grap the shaker with all the ingredients
        3. shake
        4. pour-shaker-to-shot

Here is an example of making a cocktail with ingredient 2 and ingredient 1 in shot3:
-> Filling ingredient 2
left grasp shot3
fill-shot shot3 ingredient2 left right dispenser2
pour-shot-to-clean-shaker shot3 ingredient2 shaker1 left l0 l1
clean-shot glass shot3 with ingredient2 with hand left holding shot glass and right

-> Filling ingredient 1
fill-shot shot3 ingredient1 left right dispenser1
pour-shot-to-used-shaker shot3 ingredient1 shaker1 left l1 l2
clean-shot glass shot3 with ingredient1 with hand left holding shot glass and right

-> Shake and serve
left leave shot3
right grasp shaker1
shake cocktail3 ingredient2 ingredient1 shaker1 right left
pour-shaker-to-shot cocktail3 shot3 right shaker1 l2 l1
Here is subgoal guidance for your current task, they are NOT EXACT commands, they are just guidance:
After you complete a subgoal, leave any objects you are holding.
\end{lstlisting}
\caption{PDDL Symbolic System Prompt}
\label{app:tab:pddl_symbolic}
\end{table*}

\subsection{Reflection Prompts}
Table \ref{app:all_omniref} present reflection prompts used for generating Abstract, Error, and Progress level reflections for ALFWorld, BabyAI, and PDDL datasets.
They share similar content, with the exception of examples used to demonstrate reflection examples.
In Neuro-Symbolic case, we use templated responses that are generated by OmniReflect-Symbolic on the calibration set, as few shot examples.
Table \ref{app:summarization_prompt} provides the simple prompt used for summarization across datasets, and for all models.

\begin{table*}
    
    \centering
    {
\lstset{
    backgroundcolor=\color[RGB]{255,255,255},
    breaklines=true,
    breakindent=0pt,
    basicstyle=\ttfamily\small,
    frame=trbl,
    numbers=none,
    frameround = tttt
}\begin{lstlisting}
>> Abstract
Generate a constitution specific for solving a {tasktype} task and about the environment.
The constitution should be solely based on the observation in this environment, and should not contain general rules about regular world.
The rules in the constitution should be generalizable, abstract, correct, and profound.
Some examples could include: Use microwave for heating or Tomatoes can be found in fridge, among others. <- ALFWorld
Some examples could include: If you are facing a wall, turn around and continue exploration. <- BabyAI
Some examples could include: If you have only one arm, you cannot pick up two items <- PDDL
The constitution should be in a python list format (enclosed in [])

>> Error
Generate a constitution specific for solving this task covering the potential mistakes performed so far and your suggestions on how to fix it.
The constitution should be solely based on the observation in this environment, and should not contain general rules about regular world.
The constitution should be in a python list of dictionaries format without any extra text in a single line.
You should thoroughly analyze the current trajectory and only provide feedback if a mistake happened so far. Sometimes mistakes can be indicated by the observation `Nothing happens`. 
DO NOT predict future mistakes, or share advice about future steps.
If there are no mistakes so far, then return an empty list
If efficiency of the trajectory can be improved, you should add that as well.
Here is an example: [{'mistake': 'Cabinet was not opened', 'solution': 'Open the cabinet next time}, ...] <- ALFWorld
Here is an example: [{'mistake': 'Going in circles', 'solution': 'Stop turning same way and going in circles...}, ...] <- BabyAI
Here is an example: [{'mistake': 'Attempted to pick up a block that is stacked', 'solution': 'Should use unstack...}, ...] <- PDDL
>> Progress

Critically examine the trajectory so far to solve the task, and generate explicit feedback for solving leftover subtasks.
Example: For a task of placing a heated apple in a garbage, one feedback example could be `You have heated the apple, now you should pick it up and go to garbagecan` <- ALFWorld
Example: For a task of going through a green door, one feedback example could be `You have located a green key, now pick it up and locate a green door.` <- BabyAI
Example: An example could be: I have poured ingredient 1 into the shaker. I should then shake and serve in a clean shot class. <- PDDL
The constitution should be in a python list format (enclosed in []) without any extra text in a single line.

\end{lstlisting}
}  
\caption{Reflection Prompts}
\label{app:all_omniref}
\end{table*}

\begin{table*}
    
    \centering
    {
\lstset{
    backgroundcolor=\color[RGB]{255,255,255},
    breaklines=true,
    breakindent=0pt,
    basicstyle=\ttfamily\small,
    frame=trbl,
    numbers=none,
    frameround = tttt
}\begin{lstlisting}
Inspect and summarize the constitution you have build over time by exploring the environment and solving numerous tasks.
The resulting summary should be usable by any other agent to quickly solve tasks by using the knowledge built using your experience. <- Abstract
The resulting summary should be usable by any other agent to avoid making any mistakes that were made. <- Error
There should not be duplicates in the constitution. You should be clear and concise while summarizing.
You can create new rules by summarizing multiple rules together without losing information.
Here is the current constitution: [...]
The summarized constitution should be in a python list format (enclosed in []).
\end{lstlisting}
}  
\caption{Summarization Prompt}
\label{app:summarization_prompt}
\end{table*}

\section{Symbolic Reflections}
\label{app:symbolic_reflections}
We use engineered prompts and regular expressions for generating symbolic reflections.
All prompts have been presented in Section \ref{app:prompts}.
The regular expressions used for progress tracking in OmniReflect-Symbolic system are presented below.
\noindent\textbf{ALFWorld.}
Example of ALFWorld tracking and reflection regular expressions we used can be found in Table \ref{app:ex:progress_tracking}.
\begin{table*}
\lstset{
    backgroundcolor=\color[RGB]{255,255,255},
    breaklines=true,
    breakindent=0pt,
    basicstyle=\ttfamily\small,
    frame=trbl,
    frameround = tttt,
    numbers=none,
}\begin{lstlisting}
Type: examine
Goal: look at bowl under the desklamp.
Patterns:
^(?=.* you see)(?=.*a bowl \d+)
You pick up the bowl \d+
^(?=.* you see)(?=.*a desklamp)
--------

Type: puttwo
Goal: put two soapbar in garbagecan.
Patterns:
^(?=.* you see)(?=.*a soapbar \d+)
You pick up the soapbar \d+
You put the soapbar \d+ in/on the garbagecan \d+
^(?=.* you see)(?=.*a soapbar \d+)
You pick up the soapbar \d+
You put the soapbar \d+ in/on the garbagecan \d+
--------

Type: cool
Goal: put a cool tomato in microwave.
Patterns:
^(?=.* you see)(?=.*a tomato \d+)
You pick up the tomato \d+
You cool the tomato \d+ using the
You put the tomato \d+ in/on the microwave \d+
--------
\end{lstlisting}
\caption{Progress Tracking Regular Expressions Examples}
\label{app:ex:progress_tracking}
\end{table*}

\noindent\textbf{BabyAI.}
We provide reflection for these situations:
\begin{itemize}
    \item You are going in a circle, just turn right or turn left and move forward and check valid actions.
    \item You have found a key, use it to open a same colored locked door in the path if needed. DO NOT DROP the key before you unlock the necessary doors.
    \item I have found the door, If my task is to unlock a door, I will unlock the door, else I will toggle and go through the door. template for this action is toggle and go through <color> [closed|locked] <door> <id>
    \item I found the target object. I will move towards it and pick it up if needed.
    \item Now you should drop this in a free location that does not block the path for my next steps. You cannot carry two items, so you MUST drop this before picking up the next item. DO NOT DROP if you holding a KEY. KEY should be used to unlock the door and then you can drop it.
\end{itemize}

\noindent\textbf{PDDL.}
Apart from sharing the left-over subgoals, heuristic reflection takes a simple form of `You are doing: ' and `you should be doing: ' where information is populated under the following situations:
\begin{itemize}
    \item Gripper
    \begin{itemize}
        \item When objects that are at the destination are being accidentally picked up again
        \item When the agent is performing sub-optimal pick ups
        \item When the item incorrectly drops an object
    \end{itemize}    
    \item Blockworld
        \begin{itemize}
        \item When agent attempts to use incorrect commands, such as pick up for unstack
        \item When the agent is performing sub-optimal stacking, i.e. if the goal state is 1,2,3,4, it attempts to perform 1,2 and 3,4 separately. Since stacks cannot be stacked on top of each other, we warn the agent when it enters this situation
        \item When the agent is stuck in a loop of stacking and unstacking incorrect blocks
    \end{itemize}    
    \item Barman
        \begin{itemize}
        \item When agent does not leave objects in between tasks, which leads to incorrect grasping
        \item When agent uses unclean objects, and does not perform cleaning, for instance, pour-shot-to-clean-shaker does not do anything if a clean shaker is unavailable
        \item When the agent incorrectly assumes task is done, due to incorrect order of ingredient mixing
        \end{itemize}    
    \item Tyreworld
        \begin{itemize}
        \item When agent retrieves unnecessary tools
        \item When agent forgets steps required for preparation such as removing nuts or jacking a wheel
    \end{itemize}    
\end{itemize}
\end{document}